# scMIR: a vision–language foundation model for single-cell light microscopy image representation

Yifan Shang[1, 2], Jiahui Tan[2], Xiangxiang Zeng[2, *] & Renjie Zhou[1, *]

[1] Laser Metrology and Biomedicine Lab, Department of Biomedical Engineering, The Chinese University of Hong Kong, Hong Kong, China

[2] State Key Laboratory of Chemo and Biosensing, College of Computer Science and Electronic Engineering, Hunan University, Changsha 410082, China

*e-mail: xzeng@hnu.edu.cn; rjzhou@cuhk.edu.hk

**Abstract**: Single-cell light microscopy images have become an important data source for characterizing cell phenotypes, but their complexity and heterogeneity pose challenges to high-throughput automated analysis. Existing representation learning methods mostly rely on task-oriented modeling, which is limited by specific datasets and predefined tasks, making them difficult to generalize across different cell types and microscopy modalities, and experimental conditions. Although general-purpose methods have improved the generalization ability of image representation in recent years, their limited utilization of experimental background and biological context information still poses challenges in complex phenotypic analysis. Here, we propose scMIR, a vision-language foundation model for single-cell light microscopy image representation. By synergistically combining self-supervised image reconstruction with text-guided cross-modal alignment, scMIR can simultaneously encode morphological and biological semantic information in a unified representation space. scMIR is pre-trained on 207,957 image-text pairs, covering various cell types, microscopy modalities, and perturbation conditions. scMIR outperforms existing general models and task-oriented methods as systematically evaluated on various complex tasks using 16 benchmark datasets, including cell classification, clustering, phenotype inference, and batch effect correction tasks. Furthermore, scMIR shows a strong generalization ability across various tasks without requiring task-specific fine-tuning. With its unique advantages, we envision scMIR may promote the standardization and automation of high-throughput phenotyping workflows through supporting various downstream analysis tasks.

Keywords: Single-cell microscopy; foundation model; image representation learning; vision–language model; multimodal learning

## Introduction

Over recent years, rapid development of high-throughput imaging technologies have contributed to the accumulation of large-scale cell microscopic image data, enabling the systematic learning of the correlation between cell morphological characteristics and phenotypic state [1-3]. Single-cell light microscopy images have become an important data source for frontier research in cellular phenotyping, cancer pathology, spatial biology, and computational biomedical imaging, as well as for applications in the biotechnology and pharmaceutical industries [4-9]. However, technical variations arising from different microscopy modalities, experimental conditions, and acquisition protocols pose significant challenges for developing robust and generalizable automated analysis methods across diverse experimental settings [10-12]. Therefore, developing cell image representation methods that can be stably generalized across different experimental conditions and imaging scenarios is crucial for automated microscopic image analysis [13, 14]. Traditional representation methods, mostly relying on predefined morphological features, such as intensity, shape, texture, and granularity [15, 16], they are mostly computationally inefficient, sensitive to image quality, and insufficient to capture the complexity and diversity of cellular phenotypes, thus limiting their scalability and reliability in large, heterogeneous datasets [17-19].

In recent years, deep learning-based image representation methods have become the mainstream paradigm in microscopic image analysis, capable not only of learning complex cell morphological features, capturing subtle phenotypic differences, and resolving cell phenotypic heterogeneity, but also of mining potential patterns that are difficult for humans to recognize from large-scale image data, thus providing possibilities for discovering new biological laws and mechanisms [20-25]. Existing approaches broadly fall into two categories: task-oriented representation methods and general-purpose representation learning methods. Task-oriented approaches are typically optimized for specific downstream objectives—such as cell type classification, cell cycle prediction, or perturbation response modeling [26-32]—and are trained on relatively homogeneous datasets. While effective within their intended scope, these models often generalize poorly across datasets, tasks, or microscopy types. In contrast, general-purpose methods are seeking to learn transferable representations by pretraining on diverse image collections, including using natural image datasets (e.g. ImageNet [33]) and microscopy-specific pretrained models (e.g. CytoImageNet [34] and Microsnoop [35]), which have demonstrated competitive performance across multiple downstream tasks. However, applying existing general methods for microscopic image representation face challenge: cell phenotypes not only reflect the intrinsic biological state of cells but are also influenced by multiple factors such as microscopy modalities, experimental conditions, and data acquisition processes. Therefore, learning cell representations based solely on visual information may fail to fully characterize differences in cell states within complex experimental contexts, especially in analysis tasks across datasets, experimental conditions, and imaging scenarios, where they are easily affected by technical variations and experimental biases. In contrast, incorporating contextual information such as experimental conditions, cell type, and biological perturbations, and employing visual-language multimodal joint modeling, holds promise for learning more robust cell representations.

Visual-language models have demonstrated powerful semantic representation capabilities and cross-scene generalization abilities in recent years in fields such as natural image analysis and digital pathology [36-41]. By aligning visual features with semantic information, language acts as a semantic scaffold that connects heterogeneous data sources, enhancing generalization and enabling context-aware reasoning [42, 43]. In the natural image domain, models such as CLIP, CoCa and BLIP-2 [44-46] learn transferable, semantically enriched representations from large-scale image–text pairs; similarly, in digital pathology, methods such as PLIP, CONCH and MUSK [47-49] integrate visual and semantic cues by combining tissue images with text information, demonstrating strong potential in clinically relevant tasks. Furthermore, in single-cell light microscopy images, metadata such as experimental conditions, perturbation information and imaging configurations naturally provide structured semantic descriptions of cellular states, offering a foundation for joint modeling of vision and semantics. These factors suggest that incorporating semantic information holds promise for improving representation learning of complex phenotypes and enhancing generalization capabilities across datasets and conditions.

In this paper, we propose scMIR, a vision–language foundation model for single-cell microscopy image representation. scMIR is first pre-trained on 207,957 image–text pairs spanning diverse cell types, microscopy modalities and perturbation conditions. Then, structured experimental metadata are used to provide semantic supervision, enabling explicit modeling of the relationship between cellular states and experimental conditions. Furthermore, scMIR is jointly optimized cross-modal alignment and self-supervised image reconstruction, thus enabling unified encoding of cellular morphology and biological semantics to improve characterization of complex phenotypic variation. We systematically evaluated scMIR across 16 benchmark datasets, including tasks such as cell classification, clustering, phenotypic inference and batch effect correction. Without task-specific fine-tuning, scMIR outperforms existing general-purpose [34, 35, 50] and task-oriented methods across tasks and experimental conditions, demonstrating strong generalization. Notably, across 17 classification tasks, scMIR achieved an average performance improvement of 23.95% over task-oriented methods and at least 9.2% over existing general-purpose methods. These results highlight the potential of multimodal representation learning to disentangle biological variation from technical effects, providing a robust foundation for scalable phenotypic analysis across diverse microscopy datasets.

## Results

### Overview of framework

scMIR is a vision–language foundation model designed to learn generalizable and transferable representations from diverse single-cell light microscopy data (Fig. 1). The framework consists of three main components: large-scale dataset curation, a multimodal representation learning architecture, and downstream tasks evaluation. We first curated and harmonized publicly available microscopy datasets spanning multiple species, imaging modalities, cell types, and perturbations to capture broad biological and experimental diversity (Fig. 1a; dataset details are provided in the Supplementary information). To provide consistent biological and experimental context, structured textual descriptions were constructed using controlled keywords, including species, data source, microscopy type, cell type, and related annotations (Fig. 1b). scMIR integrates complementary learning objectives to capture both semantic and morphological information. A vision–language alignment module associates images with structured textual descriptions to enable cross-dataset semantic consistency, while an image reconstruction module preserves fine-grained morphological features to stabilize representation learning (Fig. 1b).

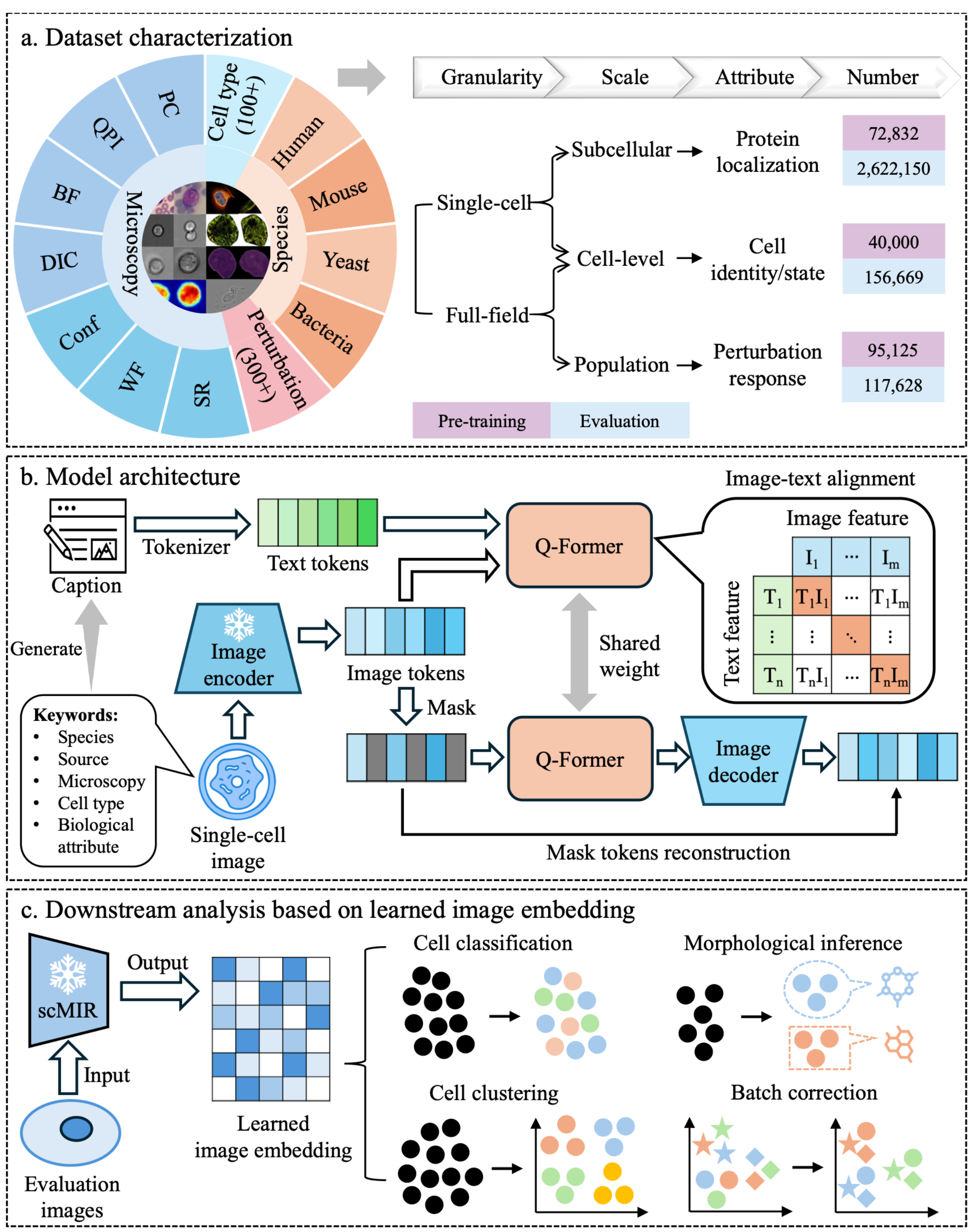

**Fig. 1** | Overview of framework. **a**. We curated a diverse collection of light microscopy datasets. The left panel summarizes dataset diversity across microscopy types, cell types, perturbation conditions and species. The right panel presents a hierarchical organization of the datasets, grouped into pre-training and evaluation sets, with the number of images annotated at each level of the hierarchy. Microscopy types are

defined using broad categories; in particular, SR datasets are fluorescence-based and are therefore grouped together with WF and Conf microscopy, without further subclassification (PC: phase contrast; QPI: quantitative phase imaging; BF: bright-field; DIC: differential interference contrast; Conf: confocal; WF: widefield fluorescence; SR: super-resolution). **b**. scMIR integrates visual and semantic supervision for single-cell microscopy image representation learning. Structured keyword annotations (for example, species, cell source, microscopy type, cell type and biological attributes) are extracted to construct concise and standardized textual descriptions. A frozen image encoder is coupled with a Q-Former initialized from BLIP-2 [44] and further optimized using image–text pairs through a vision–language alignment objective, in which structured text serves as a semantic anchor. In parallel, an image reconstruction objective operates on masked image features to preserve fine-grained morphological information. These complementary objectives jointly guide representation learning. **c**. The pretrained scMIR model is applied to the evaluation datasets to extract image embeddings, which are used for a wide range of downstream analysis tasks.

To evaluate the generalizability and robustness of the learned representations, the pretrained scMIR model was applied to a diverse set of downstream tasks using evaluation datasets (Fig. 1c). These tasks include supervised cell classification, unsupervised cell clustering across heterogeneous datasets, morphology-informed phenotypic inference under perturbations and protein localization settings, and batch correction on datasets with pronounced technical variation. These evaluations provide a comprehensive assessment of scMIR's representation quality and transferability.

**scMIR enables robust cell classification across diverse microscopy datasets**

To evaluate the robustness and generalization of scMIR for classification, we benchmarked it against four representative baselines spanning both task-specific and general-purpose representation learning paradigms: task-oriented end-to-end models originally proposed for each dataset, EfficientNetB0 [50] pretrained on natural images, and two microscopy-specific pretrained models, CytoImageNet [34] and Microsnoop [35]. We conducted a comprehensive evaluation on 13 single-cell microscopy datasets comprising 17 classification tasks, covering four species and seven microscopy types. These tasks include cell type, cell cycle, and cellular state classification, as well as protein localization and drug perturbation prediction.

Task-oriented baselines were either reproduced or directly adopted from the original studies and trained end-to-end using dataset-specific classification heads (model details are provided in the Supplementary information). In contrast, scMIR and other pretrained models were evaluated using a two-stage strategy, in which the frozen image encoder was first used to extract features, followed by training a lightweight classifier consisting of a single hidden layer and an output layer. Notably, scMIR operates in an image-only setting during downstream evaluation, without access to textual inputs, ensuring a fair comparison with vision-only baselines. All methods used identical training and test splits, and performance was assessed using the F1 score. To reflect practical scenarios in which annotated single-cell microscopy data are limited, all evaluations were conducted under constrained training set sizes, emphasizing representation quality rather than task-specific model capacity.

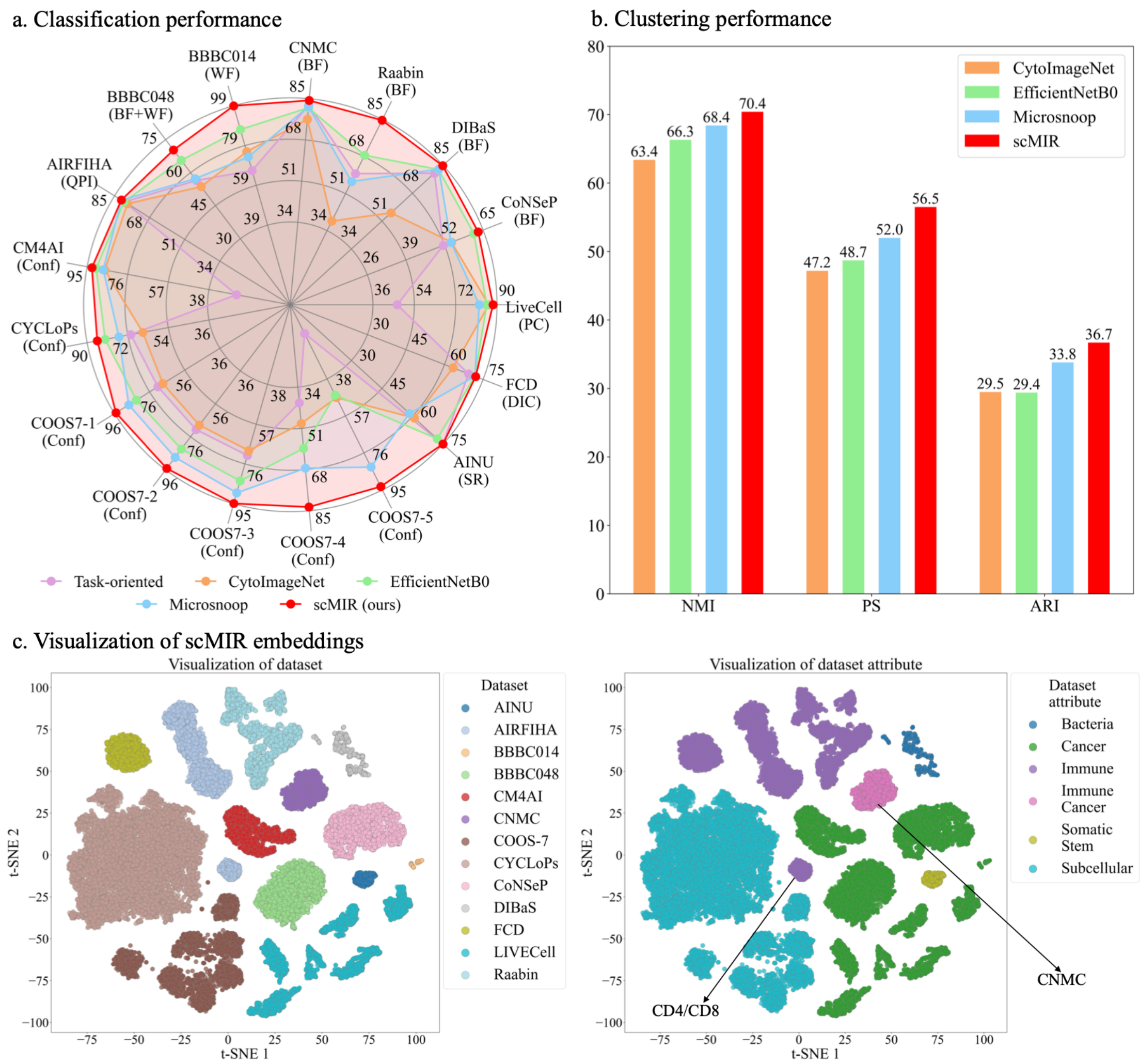

**Fig. 2** | Classification and clustering performance, visualization of scMIR embeddings. **a**. Radar plot of F1-score performance (scaled ×100) across 17 classification tasks from 13 light microscopy datasets. scMIR is compared with task-oriented models, CytoImageNet, EfficientNetB0, and Microsnoop. **b**. Quantitative assessment of unsupervised cross-dataset alignment using normalized mutual information (NMI), purity score (PS), and adjusted Rand index (ARI) computed from joint embeddings (metrics scaled ×100). **c**. t-SNE visualization of scMIR embeddings. Left, embeddings colored by dataset identity demonstrate compact dataset-specific clustering despite experimental heterogeneity. Right, the same embeddings colored by dataset-level biological attributes reveal coherent higher-level biological organization across studies and microscopy modalities. Source data are provided as a Source Data file.

Across all 17 tasks, scMIR consistently achieved the best performance (Fig. 2a), outperforming task-oriented models by an average of 23.95%, CytoImageNet by 20.07%, EfficientNetB0 by 9.20%, and Microsnoop by 9.56%. Performance gains were most pronounced on datasets with characterized by complex imaging backgrounds (BBBC014 [51], Raabin [52]), heterogeneous data (BBBC048 [31], which combines two microscopy types), and strong batch effects (COOS7 [53]). In these challenging settings, task-oriented models and vision-only pretrained representations exhibited substantial performance degradation, whereas scMIR maintained robust and stable classification performance.

Compared with task-oriented end-to-end models, scMIR reduces overfitting to dataset-specific biases by leveraging shared structure learned during large-scale pretraining. Moreover, relative to existing image-only pretrained models—whether trained on natural images or microscopy images alone—scMIR's incorporation of structured biological and experimental context during multimodal pretraining leads to more discriminative and robust visual representations, resulting in consistently improved classification performance. In summary, these findings demonstrate that scMIR provides a robust and transferable solution for cell classification across heterogeneous single-cell microscopy datasets.

**scMIR enables unsupervised semantic alignment of heterogeneous data**

While scMIR demonstrates strong generalizability in downstream classification within individual datasets, we further investigated whether image representations support unsupervised alignment across heterogeneous single-cell datasets with diverse biological and experimental contexts and microscopy types. We extracted image embeddings from all 13 datasets using scMIR and three representative pretrained baselines. The resulting embeddings were projected into a shared embedding space for joint analysis, and the global organization was quantitatively evaluated using normalized mutual information (NMI), purity score (PS), and adjusted Rand index (ARI) (Fig. 2b). Across all three metrics, scMIR consistently outperformed the baseline models, indicating superior unsupervised alignment across datasets.

We first colored scMIR embeddings by dataset (Fig. 2c left). Samples from the same dataset formed compact and coherent clusters, despite substantial variation in microscopy type, species, and acquisition conditions, demonstrating dataset-level consistency. Beyond dataset-level grouping, scMIR further organizes cell from heterogeneous datasets according to higher-level biological characteristics. When embeddings were colored by dataset attributes—such as bacterial, cancer-related, immune-related, and subcellular localization tasks—datasets sharing similar biological properties were embedded in proximity, even though they originate from different studies and microscopy types (Fig. 2c right). For example, the CNMC [29] dataset, which contains both normal and cancer-related white blood cells, was positioned between immune-related and cancer-related regions of the embedding space, reflecting its mixed biological nature. Additionally, CD4/CD8 cell populations from the AIRFIHA [54] dataset were embedded closer to cancer-related datasets, consistent with the known association between T cell subtypes and tumor-related immune responses [55-57].

These observations indicate that scMIR learns visual representations in which phenotypically similar cells are grouped according to text information learned during multimodal pretraining. Even without textual input during inference, the model effectively associates novel or heterogeneous phenotypes with these anchors, enabling robust cross-dataset organization. These results show that scMIR provides a unified embedding space that supports unsupervised semantic alignment and hierarchical structuring of heterogeneous single-cell microscopy images.

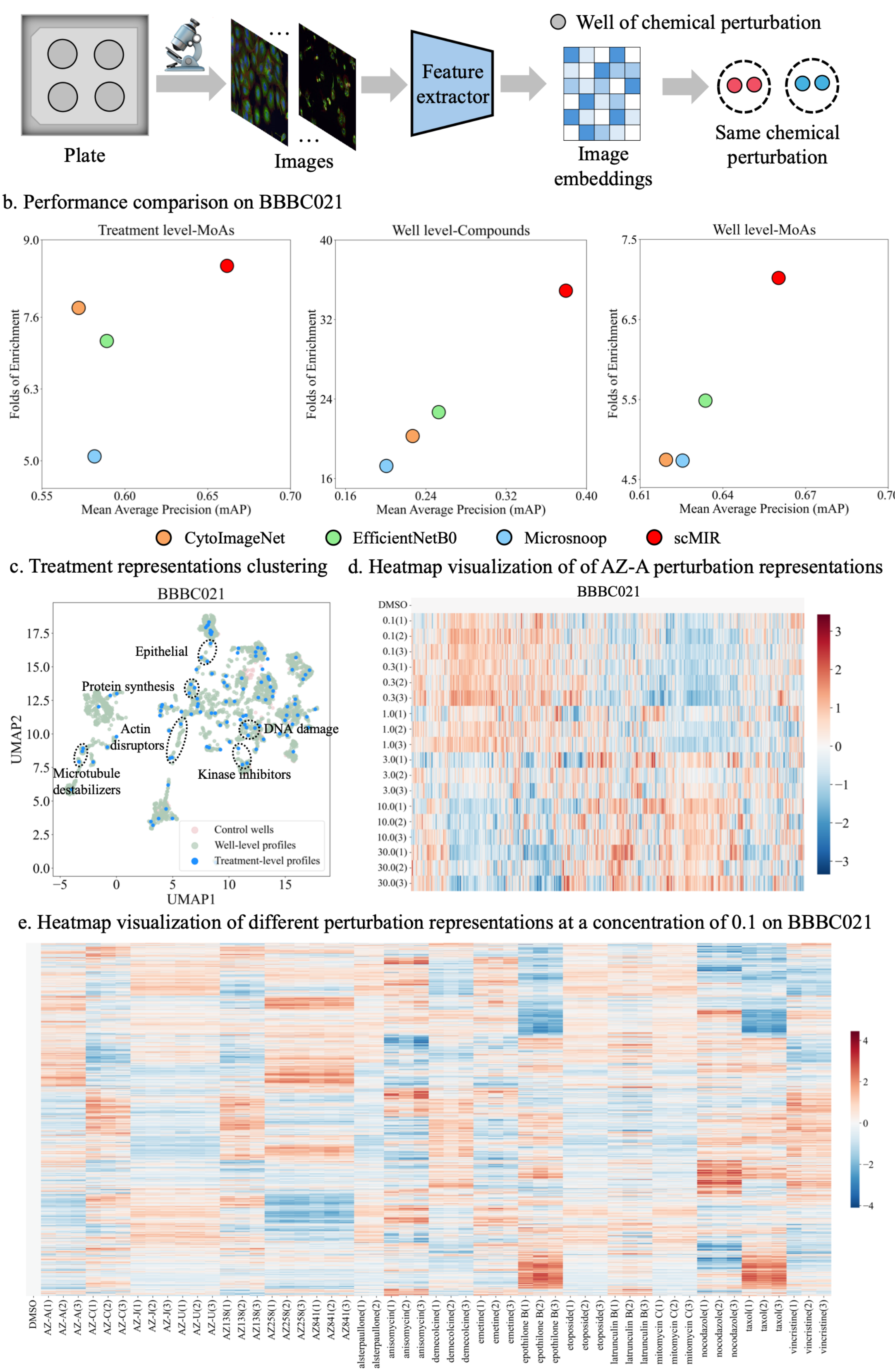

**Fig. 3** | Perturbation phenotypic profiling of scMIR embeddings. **a**. Schematic overview of phenotypic inference from perturbation image embeddings. Chemical perturbations

are applied across multi-well plates, generating microscopy images in which each well corresponds to a distinct treatment condition. Image features are extracted and used to quantify phenotypic similarity, enabling identification of identical perturbations and inference of shared mechanisms of action. **b**. Evaluation of drug-induced phenotypic similarity on BBBC021 [58]. Performance is assessed at both the well and treatment levels, capturing same perturbations and shared mechanisms of action. Scatter plots report mean average precision (mAP; x axis) versus folds of enrichment (y axis) for different feature extractors. **c**. UMAP projection using well-level and treatment-level features provided by scMIR. Drugs in black circles have the same mechanism of action. **d**. Heatmap visualization of the well-level features of AZ-A at different concentrations shows that the different concentration treatment groups have changed significantly compared with the DMSO controls. **e**. Heatmap visualization of well-level features demonstrates significant feature shifts in drug-treated groups compared to DMSO controls. Source data are provided as a Source Data file.

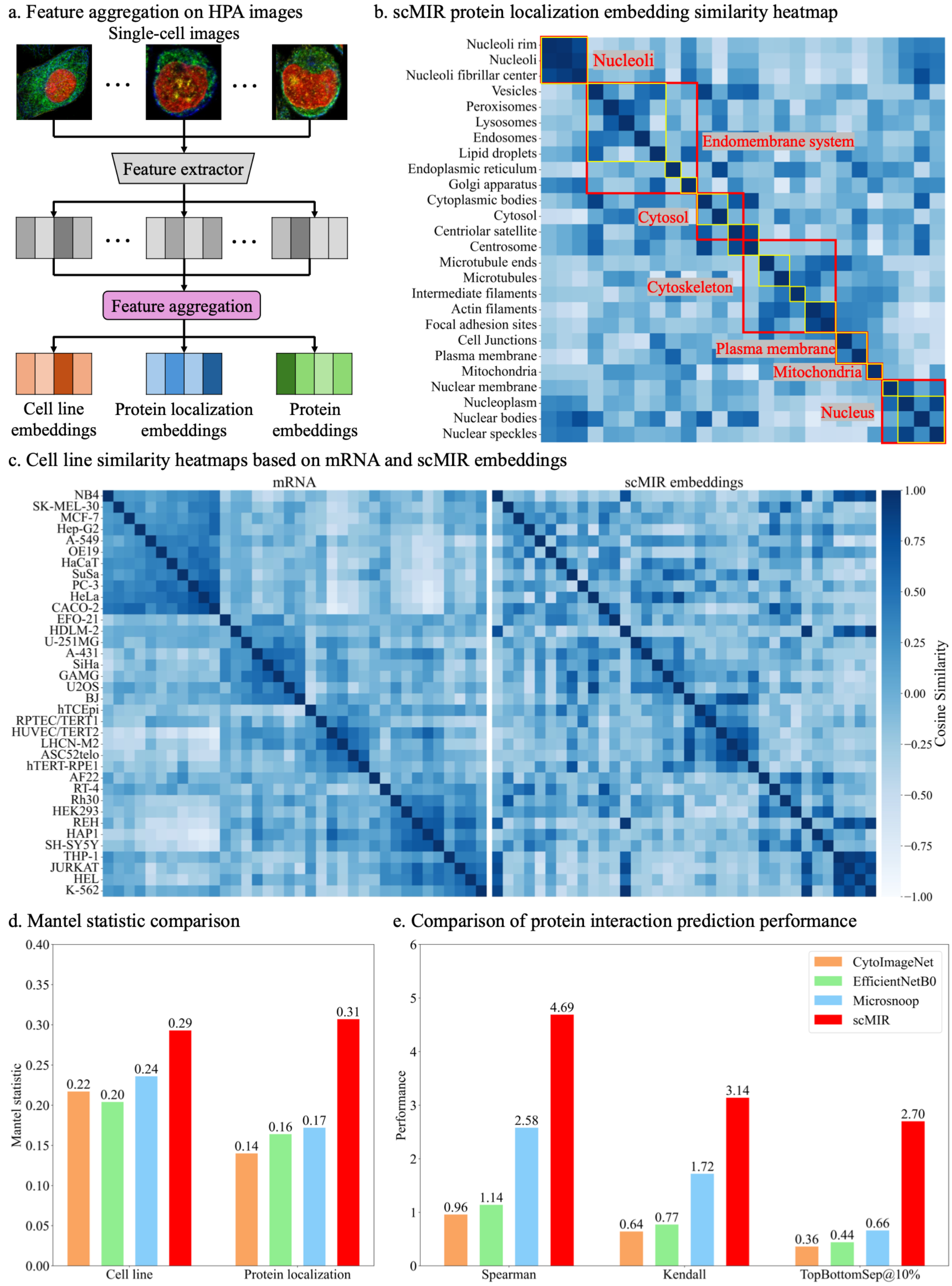

**Fig. 4** | **a**. Schematic illustration of hierarchical feature aggregation on the Human Protein Atlas (HPA) dataset. Single-cell image embeddings are first extracted from individual microscopy images and then aggregated to obtain higher-level representations at multiple biological scales, including cell line–level, protein localization–level, and protein-level features. **b**. Similarity matrix of protein localization groups computed from scMIR-derived protein-level features, with expert-defined hierarchical organelle annotations indicated (red and yellow). **c**. Cell-line similarity matrices derived from bulk mRNA expression (left) and morphology-

informed profiles aggregated from scMIR single-cell embeddings (right), showing partial correspondence between transcriptional and morphological organization. **d**. Mantel statistics comparing morphology-informed profiles with bulk mRNA profiles and expert-defined localization hierarchies across four methods. **e**. Protein–protein interaction prediction based on protein-level morphological representations, evaluated against STRING using Spearman and Kendall correlations and TopBottomSep@10% (scaled ×100). Source data are provided as a Source Data file.

**scMIR encodes biologically relevant information**

To assess whether scMIR embeddings encode biologically meaningful information beyond visual discriminability, we examined their ability to organize cellular phenotypes in drug perturbation and protein localization settings.

We evaluated scMIR on the Cell Painting [3] drug perturbation dataset, which consists of multi-channel fluorescence full-field images acquired under diverse chemical treatments. Single cells were segmented using Cellpose [59], embedded with scMIR, and hierarchically aggregated to derive site-, well-, and treatment-level representations. Phenotypic similarity was assessed based on the consistency of embeddings across shared drug perturbations and mechanisms of action (Fig. 3a). On BBBC021 [58], scMIR outperformed three general baselines in grouping phenotypically similar treatments (Fig. 3b), indicating enhanced sensitivity to drug-induced morphological variation. We further utilize UMAP to perform visual analysis on the well-level and treatment-level representations of the BBBC021. The results show (Fig. 3c) that scMIR can spontaneously form clusters according to the compound's mechanism of action (MoA). Compounds with the same MoA show higher proximity in the representation space, while compounds with different MoA are clearly separated, indicating that the model can extract discriminative features related to biological functions from cell morphological phenotypes. Furthermore, as shown in Fig. 3d–e, compared with the DMSO control, each drug treatment group produced significant characterization shifts; repeated wells under the same treatment conditions remained highly consistent, while clear distinctions were shown between different concentrations and different drug categories, indicating that scMIR can stably capture drug-induced phenotypic changes and effectively characterize dose-dependent and mechanism-related biological differences.

We next applied scMIR to the Human Protein Atlas (HPA) dataset [12] to evaluate whether morphology-informed embeddings capture higher-level biological organization (Fig. 4a). Clustering based on protein localization embeddings further recovered subsets of the hierarchical organelle groupings defined by expert annotations (Fig. 4b). Single-cell embeddings were aggregated to construct morphology-informed profiles for 36 cell lines. Clustering of these profiles revealed multiple modules. These modules partially overlapped with similarity patterns derived from bulk mRNA expression data (Fig. 4c). We performed quantitative evaluation by computing Mantel statistics between embedding-derived similarity matrices and corresponding biological reference matrices. Across all comparisons, scMIR consistently achieved the strongest agreement among evaluated methods (Fig. 4d). In addition, protein-level morphological representations constructed from scMIR embeddings showed the highest correlation with protein–protein interaction scores obtained from STRING [60] (Fig. 4e).

These results demonstrate that scMIR learns morphology-informed single-cell representations that reflect underlying biological structure across multiple levels of organization. Although no textual or contextual input is used during downstream inference, semantic guidance introduced during multimodal pretraining enables scMIR to encode biologically relevant phenotypic information directly into visual representations.

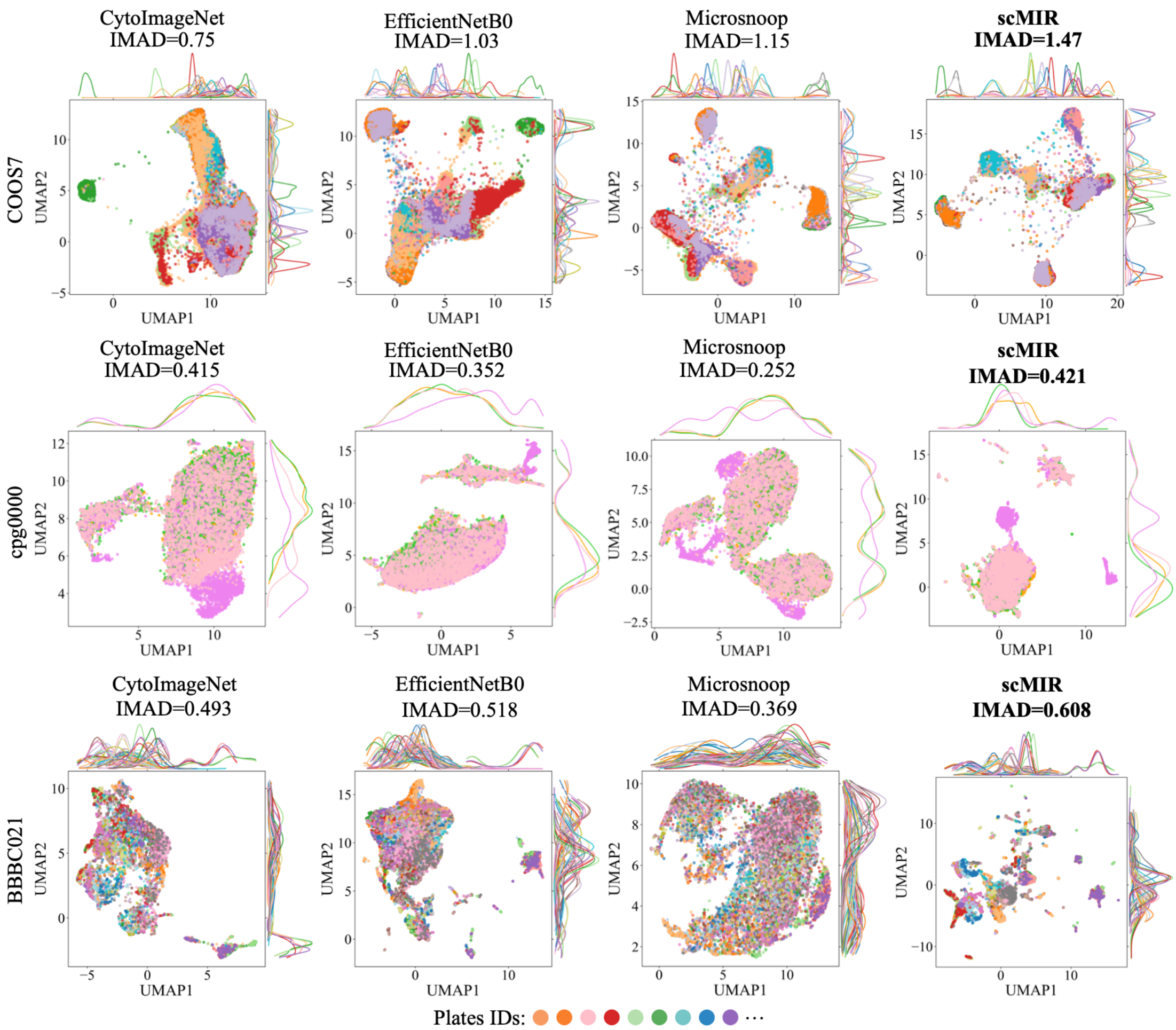

**Fig. 5** | Batch effect of image embeddings. UMAP visualizations of image-level features extracted by scMIR and representative pretrained baselines on the COOS7 [53], cpg0000 [1] and BBBC021 [58] datasets. Features are colored by plate IDs. IMAD quantifies the consistency of UMAP patterns across batches, with higher values indicating improved batch-effect mitigation. UMAP: Uniform Manifold Approximation and Projection; IMAD: inverse median absolute deviation.

**scMIR effectively reduces batch effects for robust image representations**

Technical and instrumental variability can introduce batch effects across imaging plates,

confounding genuine biological phenotypes. We therefore assessed whether scMIR can disentangle biological variation from technical noise by evaluating its ability to mitigate batch effects in microscopy images.

Batch correction was performed on a single-cell microscopy dataset (COOS7 [53]) and two full-field drug-perturbation imaging datasets (cpg0000 [1] and BBBC021 [58]) by extracting image-level representations with scMIR and representative pretrained baselines, followed by joint embedding analysis across experimental plates. Batch-related variability was quantified using the inverse median absolute deviation (IMAD), with higher values indicating reduced dispersion of representations across batches. As shown in Fig. 5, several baseline methods exhibited pronounced plate-specific clustering, indicative of substantial batch effects. In contrast, scMIR markedly reduced plate-driven separation, while preserving biologically meaningful class structure. Consistently, scMIR achieved the highest IMAD scores across three datasets, demonstrating enhanced robustness to batch-induced variation.

These results indicate that scMIR learns image representations that effectively suppress technical and experimental noise while retaining biologically faithful phenotypic signals. Notably, this robustness arises without explicit batch annotations or post hoc correction, emerging instead as an intrinsic property of multimodal pretraining guided by structured biological context. Collectively, these findings underscore scMIR's capacity to produce robust, generalizable, and biologically meaningful representations for large-scale single-cell microscopy data.

## Discussion

In this study, we introduced scMIR, a vision–language foundation model for learning unified and generalizable representations from single-cell light microscopy images. We collected approximately three million images from publicly available datasets, covering a wide range of species, microscopy modalities, cell types, and experimental perturbations, for multimodal pre-training and downstream evaluation. By integrating visual information with structured experimental and biological context, scMIR enables learning of representations that generalize across heterogeneous imaging conditions. Across a wide range of downstream tasks—including cell classification, unsupervised dataset alignment, morphology-informed phenotypic inference, and batch correction—scMIR consistently outperformed task-oriented and existing general-purpose pretrained methods.

A key challenge in single-cell image analysis is separating biologically meaningful variation from technical and experimental confounders [61, 62]. scMIR addresses this challenge through multimodal pretraining with semantic guidance. Vision–language alignment links diverse cellular morphologies to shared experimental and biological descriptors, while an image reconstruction objective preserves fine-grained morphological information. These objectives enable scMIR to capture phenotypic patterns at multiple levels, from subcellular structures to cell-type- and perturbation-associated morphology. Notably, this is achieved without requiring task-specific supervision. scMIR representations recover biologically relevant structure across diverse analytical settings. In unsupervised analyses, embeddings align datasets according to shared biological properties rather than microscopy type or dataset origin. In morphology-informed phenotypic inference, scMIR-derived similarities partially correspond to transcriptional profiles, expert-curated subcellular localization categories, and known protein–protein interaction patterns. This partial correspondence is expected. Morphology reflects downstream and integrated cellular states shaped by protein organization and cellular architecture, whereas transcriptional measurements directly capture gene regulatory activity. Consequently, morphology-informed representations capture stable and functionally relevant phenotypes but are less sensitive to molecular changes that do not produce observable structural differences. In addition, scMIR shows strong robustness to batch effects, reducing technical variation while preserving biological structure. This robustness arises from semantic guidance during pretraining, which encourages the model to focus on biological features that are consistent across experimental conditions. As a result, scMIR learns representations that are more stable across datasets and imaging setups.

Despite its good generalization ability, scMIR still has several limitations. First, scMIR relies on predefined structured text annotations, which typically only cover limited experimental and biological background information, making it difficult to comprehensively describe complex cell states, experimental designs, and dynamic biological processes. Therefore, the model's representational ability may be limited when resolving more refined cell states or complex biological processes. Second, current pre-training is mainly based on static two-dimensional single-cell images, and it has not yet explicitly modeled the dynamic behavior of cells over time, nor does it depict the spatial relationships between cells and the organizational structure of populations. Therefore, its support for dynamic phenotypic evolution, cell state transitions, and tissue-level biological processes remains limited. Furthermore, although joint visual-language modeling improves the generalization ability of representations, the correspondence between current learned representations and specific biological mechanisms remains unclear, and the model still lacks deeper mechanistic explanation capabilities.

Future research can further expand scMIR in several directions. On the one hand, large-scale language models can be combined to construct more open and richer biological semantic descriptions, enhancing the model's ability to understand complex experimental backgrounds, biological processes, and domain knowledge [63, 64]. On the other hand, time-series microscopic images, three-dimensional microscopic imaging, and spatial cell tissue information can be further introduced to promote the development of models from static morphological representation to dynamic cell state modeling [65, 66]. Furthermore, combining multimodal data such as transcriptomics and proteomics is expected to establish a more systematic correlation between cell morphology and molecular state, thus laying the foundation for constructing predictable and simulable virtual cell models [67, 68].

## Methods

## scMIR model

scMIR is a pretrained model for learning transferable representations from single-cell microscopy images. During pre-training, scMIR is trained on paired single-cell images and structured textual captions to enable text-guided visual representation learning. In downstream applications, the model takes only single-cell microscopy images as input and outputs task-agnostic image embeddings. scMIR framework includes three key modules: an image encoder, an image–text alignment module, and an image reconstruction module.

## Image encoder

scMIR builds upon the BLIP-2 [44] framework and employs a frozen visual backbone to extract features from single-cell microscopy images. Specifically, we use the ViT-g/14 vision transformer from EVA-CLIP [69] as the image encoder. Freezing the image encoder allows scMIR to leverage pretrained visual priors, reduce computational cost, and limit the number of trainable parameters, facilitating efficient adaptation to single-cell microscopy image–text pairs. The extracted visual embeddings form the foundation for subsequent cross-modal representation learning.

## Image–text alignment module

The image–text alignment module maps visual features extracted from single-cell microscopy images into a shared semantic space guided by textual captions, enabling cross-modal feature learning. The following sections provide a detailed description of its key components. We first introduce the Q-Former, a query-based transformer that interacts with visual features to extract semantically meaningful embeddings. We then describe the image–text alignment objective, which supervises the model to align visual representations with textual captions, ensuring robust multimodal feature learning.

## Q-Former

We adopt the Q-Former module initialized from pre-trained BLIP-2 [44] weights to efficiently bridge the frozen image encoder and downstream multimodal representation learning. By building on a pre-trained Q-Former, we leverage strong prior knowledge from large-scale vision–language pre-training, enabling the model to extract semantically meaningful visual embeddings from single-cell microscopy images with limited domain-specific data.

The Q-Former is a trainable query-based transformer that maps high-dimensional visual features into a compact set of embeddings suitable for alignment with textual descriptions. In our implementation, we use 32 learnable query embeddings, each with a hidden dimension of 768. Queries interact through self-attention and attend to the frozen image features via cross-attention, selectively extracting the most text-relevant visual information. The module consists of two transformer subcomponents sharing the same self-attention layers: an image transformer that processes visual features from the frozen encoder, and a text transformer capable of encoding or decoding textual information. The compact query outputs ($32 \times 768$) act as a bottleneck, focusing representation capacity on semantically informative visual content while reducing computational cost.

During pre-training, the Q-Former is trained on single-cell image–text pairs to produce visual embeddings that are guided by textual semantics. The query embeddings generated by Q-Former serve as the input for the image–text alignment objective, which supervises the model to align visual and textual representations for downstream single-cell feature learning.

**Image–text alignment objective**

The image–text alignment objective supervises the Q-Former to produce visual embeddings that are semantically consistent with corresponding textual descriptions. By aligning single-cell image features with structured textual annotations, this objective enables scMIR to learn cross-modal representations that are both biologically informative and compatible with downstream tasks, including clustering, classification, and aggregation of single-cell profiles.

Follow BLIP-2 [44], we adopt three complementary pre-training objectives, each imposing a distinct type of constraint on the interaction between image and text representations:

(1) Image–Text Contrastive Learning (ITC)

ITC encourages alignment between image and text representations by maximizing their mutual similarity while contrasting with negative pairs. Formally, let $Z \in \mathbb{R}^{N_d \times d}$ denote the query-based output embeddings from the Q-Former and $t \in \mathbb{R}^{d}$ denote the text embedding corresponding to the caption. The image–text similarity is computed as:

$$s(Z, t) = Max_{i=1,\dots,N_q} \frac{Z_i \cdot t}{\|Z_i\|\|t\|}, \tag{1}$$

where $Z_i$ is the $i$-th query embedding. The contrastive loss is defined over a batch of image–text pairs as:

$$\mathcal{L}_{ITC} = -\frac{1}{B}\sum_{b=1}^{B} \log \frac{\exp\left(\frac{s(Z_b, t_b)}{\tau}\right)}{\sum_{j=1}^{B} \exp\left(\frac{s(Z_b, t_j)}{\tau}\right)}, \tag{2}$$

where $B$ is the batch size and $\tau$ is a learnable temperature parameter. A unimodal self-attention mask prevents direct interaction between queries and text during this objective, ensuring the queries extract visual information independently.

(2) Image-Grounded Text Generation (ITG)

The ITG objective trains the model to generate textual descriptions conditioned on image features. Each query embedding serves as a bottleneck to pass visual information to the text decoder, which then predicts the sequence of tokens. Let $\hat{y}_{1:T}$ denote the predicted text token probabilities and $y_{1:T}$ the ground-truth tokens. The ITG loss is defined as a cross-entropy over the text sequence:

$$\mathcal{L}_{ITG} = -\sum_{t=1}^{T} \log P(\hat{y}_t | \hat{y}_{<t}, Z), \tag{3}$$

where the queries $Z$ provide the conditioning information for text generation. Multimodal causal attention masks control the interactions, allowing queries to attend to each other and text tokens to attend to all queries and previous tokens.

(3) Image–Text Matching (ITM)

ITM is a binary classification task that determines whether an image–text pair is matched or not. For each output query embedding $Z_i$, a linear classifier produces a logit $l_i \in \mathbb{R}^2$. The matching score for the pair is obtained by averaging over all queries:

$$s_{ITM} = \frac{1}{N_q}\sum_{i=1}^{N_q} softmax(l_i)_{pos}, \tag{4}$$

where $softmax(l_i)_{pos}$ denotes the probability of the positive class. The ITM loss is computed using cross-entropy:

$$\mathcal{L}_{ITM} = -y log s_{ITM} - (1-y)\log(1-s_{ITM}), \tag{5}$$

where $y \in \{0,1\}$ indicates whether the pair is matched. Hard negative mining is applied to select informative unmatched pairs.

**Image reconstruction module**

To further regularize visual representation learning and encourage the model to capture fine-grained morphological features, we introduce an Image Reconstruction Module on top of the frozen image encoder and Q-Former. This module is designed to reconstruct masked portions of the image-level embeddings, guiding the model to retain detailed structural information in the learned visual features. By recovering masked features, the model is encouraged to encode both global context and local morphological details, which is particularly important for single-cell microscopy images.

In scMIR, a fraction of the output tokens from the frozen image encoder are randomly masked, producing a partially observed embedding sequence. These masked embeddings are fed into a shared-weight Q-Former, which operates identically to the main Q-Former used for image–text alignment, ensuring consistency in visual feature extraction. The Q-Former outputs are then projected to a lightweight Transformer decoder that predicts the original embeddings for the masked tokens. The decoder operates over a low-dimensional bottleneck, where the masked tokens are reconstructed using cross-attention to the unmasked memory from the image encoder.

Formally, let $X \in \mathbb{R}^{B\times N\times D_v}$ denote the full set of visual features extracted from a batch of images by the frozen image encoder, where $B$ is the batch size, $N$ is the number of visual tokens, and $D_v$ is the feature dimension. A binary mask $M \in \{0,1\}^{B\times N}$ is applied to randomly occlude a portion of the visual tokens, producing masked features $X_{masked} = X \odot (1-M)$. The masked features are input into the Q-Former to produce query embeddings $Z_{masked} \in \mathbb{R}^{B\times Q\times D_q}$, where $Q$ is the number of learnable queries and $D_q$ is the hidden dimension of the Q-Former.

The embeddings $Z_{masked}$ are then projected into a decoder space and passed through a lightweight transformer decoder to reconstruct the original visual features.

The reconstruction predictions are denoted as $\hat{X} \in \mathbb{R}^{B\times Q\times D_v}$. The reconstruction loss is computed only over the masked positions:

$$\mathcal{L}_{REC} = \frac{1}{|M|}\sum_{b=1}^{B}\sum_{i=1}^{N} M_{b,i} \cdot SmoothL1(\hat{X}_{b,i}, X_{b,i}), \tag{6}$$

where $|M|$ is the total number of masked tokens, and $SmoothL1(\cdot)$ denotes the Huber loss. By optimizing $\mathcal{L}_{REC}$ jointly with the image–text alignment objectives, the Q-Former is guided to extract visual features that are both semantically meaningful and morphologically detailed, improving downstream performance on single-cell feature representation and analysis tasks.

**Overall pre-training objective**

The four pre-training objectives—Image–Text Contrastive (ITC), Image-Grounded Text Generation (ITG), Image–Text Matching (ITM), and Image Reconstruction (REC)—are jointly optimized during pre-training. The total loss is:

$$\mathcal{L} = \mathcal{L}_{ITC} + \mathcal{L}_{ITG} + \mathcal{L}_{ITM} + \lambda_{REC}\mathcal{L}_{REC}, \tag{7}$$

where $\lambda_{REC}$ is a weighting factor for the reconstruction loss. This unified objective guides the model to extract visual embeddings that are semantically aligned with textual descriptions while preserving fine-grained morphological information, producing robust and generalizable representations for downstream single-cell analysis.

**Training protocol**

The model is trained in a multimodal pretraining stage using paired microscopy images and textual descriptions. During pretraining, the full scMIR architecture, consisting of the image encoder, Q-Former, and the proposed image reconstruction module, is optimized end-to-end. Detailed hyperparameter configurations and implementation settings are provided in the Supplementary information. For downstream tasks, microscopy images are fed into the pretrained model to extract fixed image-level representations, which are then used as input features for task-specific classifiers or evaluation protocols, depending on the application.

**Benchmarking methods**

We compare scMIR with both task-oriented models and general-purpose pretrained representations. For task-oriented approaches, models are trained and evaluated following the protocols described in the original studies or their validated reproductions, with all architectural and training details provided in the Supplementary information. We further benchmark against three widely used pretrained models for microscopy image representation learning, including Microsnoop [35], EfficientNetB0 [50], and CytoImageNet [34], using their official pretrained weights. For all baseline methods, fixed image-level features are extracted without additional fine-tuning and evaluated using the same downstream classifiers and evaluation protocols as scMIR to ensure fair comparison.

**Datasets**

We assemble a large-scale collection of publicly available microscopy image datasets for pretraining and evaluation of scMIR, comprising more than 3,000,000 images. The corpus includes both single-cell images and full-field microscopy images acquired under diverse experimental conditions. Full-field images are processed using automated cell segmentation tools to extract individual cell instances, which are combined with originally provided single-cell datasets to form a unified single-cell image pool. All images used in this study originate exclusively from public datasets. The resulting collection spans seven microscopy modalities, covers more than one hundred cell types, four species, and over three hundred drug or chemical perturbation conditions, capturing broad biological and experimental diversity. For multimodal pretraining, the complete single-cell image set is randomly split into pretraining and validation sets with no image-level overlap. Textual metadata are used only during pretraining and are excluded from all downstream analyses. Detailed dataset statistics and sources are provided in the Supplementary information.

## Data preprocessing

For full-field microscopy images lacking provided cell boundary annotations, single-cell instances are obtained using an automated segmentation pipeline. Specifically, we apply Cellpose [59] with the nuclear channel as the primary segmentation cue to delineate individual cells. Following segmentation, each cell region is cropped from the original image, and zero padding is applied to produce square images. All images are subsequently processed by the pretrained image encoder and Q-Former to extract visual representations. The Q-Former outputs a fixed set of 32 query embeddings for each image, each with a dimensionality of 768. To obtain a compact image-level representation, we compute the mean of the query embeddings, resulting in a single 768-dimensional feature vector per image. For multi-channel microscopy images, features are extracted independently from each channel using the same procedure. The resulting channel-specific feature vectors are then concatenated to form the final image representation, yielding a feature dimension of $n \times 768$, where $n$ denotes the number of imaging channels.

## Evaluation experiments

### Cell classification

For cell classification tasks, datasets containing full-field microscopy images are first processed by cell segmentation to obtain single-cell images. Single-cell representations are extracted using the pretrained models and then aggregated by average pooling to obtain image-level features.

For each dataset, samples are split into training and test sets. Model selection is performed on the training set using five-fold cross-validation, yielding five optimal models. These models are then evaluated on the held-out test set, and performance is reported as the average across runs. We compute ACC, F1 score, Macro-AP, and Micro-AP, and report F1 score in the main text, while the remaining metrics are provided in the source data.

To ensure fair comparison, task-oriented methods, general pretrained baselines, and scMIR are trained using the same learning rate and maximum number of epochs. For all general pretrained methods, a unified classifier architecture is adopted, consisting of a 128-dimensional hidden layer followed by a classification layer. Dataset-specific training configurations are provided in the Supplementary information.

**Cell clustering**

For unsupervised evaluation, KMeans clustering is applied to image representations extracted from all datasets. For multi-channel datasets with varying feature dimensionality, channel-wise features are first averaged to ensure a consistent feature length across datasets. Because datasets differ in sample size, we perform class-balanced subsampling for visualization, with up to 600 samples per dataset. Clustering performance is quantitatively evaluated using NMI, purity score (PS), and ARI.

**Morphology-informed phenotypic inference**

For drug-perturbation datasets, full-field microscopy images are first segmented into single-cell images, from which single-cell representations are extracted. Image-level features are obtained by average aggregation of single-cell features. Subsequently, well-level representations are computed by averaging image-level features, and treatment-level representations are derived by further aggregating wells subjected to identical perturbation conditions. To reduce the influence of technical variability and batch effects during phenotypic inference, all image-level features are subjected to sphering [70] normalization prior to feature aggregation. This preprocessing step is applied consistently to scMIR and all baseline methods, ensuring a fair comparison and improving the robustness of downstream phenotypic similarity analysis. As each treatment-level representation uniquely corresponds to a single drug, compound-level phenotypic similarity inference is not performed. Model performance is evaluated using mean Average Precision (mAP) and Folds of Enrichment (FoE).

For the Human Protein Atlas dataset [12], single-cell images are grouped according to cell line, protein localization, and protein identity. Single-cell features are aggregated to generate representations at each of these levels. Cell-line-level feature similarities are compared with similarities derived from mRNA sequencing profiles [71]. Protein-localization-level feature similarities are evaluated against expert-annotated organelle hierarchy groupings. Protein-level feature similarities are further assessed by comparing inferred protein–protein interaction scores with interaction weights provided by STRING [60]. Evaluation metrics include Mantel statistics, Spearman and Kendall correlations, and TopBottomSep@10%.

**Batch correction**

Batch correction is conducted using joint embedding visualization with UMAP [72]. Each image is treated as an individual data point. For single-cell datasets, each image corresponds directly to a single feature vector, whereas for drug perturbation datasets, image-level features are obtained by averaging single-cell representations. Batch-related variability is quantified using the Inverse Median Absolute Deviation (IMAD), enabling assessment of the extent to which learned representations suppress technical variation while preserving biological structure.

**Evaluation metrics**

For all evaluation metrics reported in this study, higher values indicate better performance.

**Accuracy (ACC)**

$$ACC = \frac{1}{N}\sum_{i=1}^{N} \mathbb{I}(\hat{y}_i = y_i), \tag{8}$$

where $N$ denotes the number of samples, $\hat{y}_i$ and $y_i$ are the ground-truth and predicted labels, respectively, and $\mathbb{I}(\cdot)$ is the indicator function.

**F1 score**

$$F1 = \frac{2\cdot Precision\cdot Recall}{Precision+Recall}, \tag{9}$$

where $Precision = \frac{TP}{TP+FP}$ and $Recall = \frac{TP}{TP+FN}$ are computed based on true positives ($TP$), false positives ($FP$), and false negatives ($FN$).

**Macro-Average Precision (Macro-AP)** is obtained by computing the average precision independently for each class and then averaging across all classes, treating each class equally regardless of sample size.

**Micro-Average Precision (Micro-AP)** computes average precision by aggregating true positives, false positives, and false negatives over all classes before calculating precision–recall statistics, thereby weighting classes by their sample frequencies.

**Normalized Mutual Information (NMI)** [73]

$$NMI = \frac{2I(\hat{Y};Y)}{H(\hat{Y})+H(Y)}, \tag{10}$$

where $\hat{Y}$ is predicted cluster assignment, $Y$ is ground-truth label, and $I(\cdot;\cdot)$ denotes mutual information and $H(\cdot)$ denotes entropy.

**Purity score (PS)** [74]

$$PS = \frac{1}{N}\sum_{k} max_j\left|C_k \cap Y_j\right|, \tag{11}$$

where $C_k$ denotes the set of samples in cluster $k$ and $Y_j$ denotes the set of samples belonging to class $j$.

**Adjusted Rand Index (ARI)** [75]

$$ARI = \frac{\sum_{ij}\binom{n_{ij}}{2}-\left[\sum_i\binom{a_i}{2}\sum_j\binom{b_j}{2}\right]/\binom{n}{2}}{\frac{1}{2}\left[\sum_i\binom{a_i}{2}+\sum_j\binom{b_j}{2}\right]-\left[\sum_i\binom{a_i}{2}\sum_j\binom{b_j}{2}\right]/\binom{n}{2}}, \tag{12}$$

where $n_{ij}$ denotes the number of samples that are assigned to clusters $i$ and $j$ based on the true labels and the clustering labels, respectively, $a_i$ is the number of samples from cluster $i$ based on the true labels and $b_j$ is the number of samples assigned to cluster $j$ according to the clustering labels.

**Mean Average Precision (mAP)** evaluates phenotypic similarity retrieval performance. For each query treatment, all other treatments are ranked by representation similarity, and those sharing the same mechanism of action or pathway are considered relevant. The average precision is computed from the precision–recall curve of the ranked list, and mAP is obtained by averaging over all query treatments.

**Folds of Enrichment (FoE)** quantifies the enrichment of biologically related treatments among top-ranked phenotypic neighbors. For each query treatment, a 2 × 2 contingency table is constructed by counting treatments with the same versus different mechanisms of action or pathways above and below a predefined similarity threshold (top 1%). A one-sided Fisher's exact test is used to compute the odds ratio, and FoE is defined as the average odds ratio across all query treatments.

**Spearman**

$$\rho = 1 - \frac{6\sum d_i^2}{n(n^2-1)}, \quad (13)$$

$$d_i = rank(y_i) - rank(\hat{y}_i), \quad (14)$$

where $n$ is the total number of samples, $y_i$ is the true value of the $i$-th sample, $\hat{y}_i$ is the predicted value of the $i$-th sample, $rank(\cdot)$ is the sorting position.

**Kendall correlation**

$$\tau = \frac{n_c - n_d}{\frac{1}{2}n(n-1)}, \quad (15)$$

where $n$ is the total number of samples, $n_c$ is the number of consistent pairs, and $n_d$ is the number of inconsistent pairs.

**TopBottomSep@10%**

$$TopBottomSep@10\% = \frac{1}{k}\sum_{i=n-k+1}^{n} \hat{y}_i - \frac{1}{k}\sum_{i=1}^{k} \hat{y}_i, \quad (16)$$

where $n$ is the total number of samples, $k = n * ratio$ is the size of the head/tail subset (ratio=0.1 in this paper), $y$ is the $i$-th true value after sorting in ascending order of true values, and $\hat{y}$ represents the predicted similarity corresponding to the $i$-th sample after sorting by true similarity. This metric measures the model's ability to predict and separate real highly similar and low-similar sample pairs; a higher value indicates a stronger ability to distinguish between them.

**Inverse Median Absolute Deviation (IMAD)** is defined as the inverse of the median absolute deviation (MAD) of feature distances measured across experimental batches. Higher IMAD values indicate reduced dispersion of representations across batches and therefore weaker batch effects.

## Data availability

The datasets generated during the current study are available from the corresponding author upon reasonable request.

## Code availability

The code is available from the corresponding author upon reasonable request.

## Acknowledgements

The authors thank L.T., T.Y. and Z.R. for assistance with model design and insightful suggestions. This work was supported by Research Grant Council of Hong Kong SAR under the Grant R4024-23, National Natural Science Foundation of China Fund (Grant Nos. 62422513, 62425204, U22A2037, 62450002, 62432011, 62402166), Hong Kong Scholar Program under the Grant XJ2024022, and Natural Science Foundation of Hunan Province under the Grant 2024JJ6158.

## Author contributions

Y.S., X.Z. and R.Z. conceived and designed the study. Y.S. performed the experiments under the supervision of X.Z. and R.Z., prepared the figures, and wrote and revised the manuscript. J.T. collected the data and contributed to data analysis; X.Z. and R.Z. supervised the project and revised the manuscript. All authors reviewed and approved the manuscript.

# scMIR: a vision–language foundation model for single-cell light Microscopy Image Representation

## Supplementary information

## MATERIALS AND METHODS

### Pre-training datasets

We curated six publicly available microscopy datasets for multimodal pretraining, including LIVECell, cpg0000, PBC, HPA, HepG2, and B_ALL. Among them, LIVECell and cpg0000 consist of full-field microscopy images, whereas the remaining datasets provide single-cell images. For LIVECell, we directly used the cell instances provided with the original dataset. For cpg0000, full-field images were segmented into single-cell instances using the Cellpose [1] framework, with the Hoechst channel serving as the nuclear reference. When segmented cell images did not conform to a square shape, zero-padding was applied to obtain square inputs. For LIVECell, cpg0000, and HPA, the original pretraining and evaluation splits provided by the datasets were preserved, ensuring no overlap between the two subsets. Detailed descriptions of each dataset, along with a comprehensive summary of the pretraining data statistics, are provided in Table S1.

**LIVECell** [2]

LIVECell is a large-scale, high-quality phase-contrast microscopy dataset with expert-validated annotations, comprising 5,239 images and over 1.6 million annotated cells spanning diverse cell morphologies and culture densities. The dataset includes eight cell types, covering seven human cell lines (A172, BT-474, Huh7, MCF7, SH-SY5Y, SkBr3, and SK-OV-3) and one mouse cell line (BV-2). In this study, we selected 351 images to generate single-cell instances for pretraining using the provided segmentation annotations, while the remaining images were reserved as an evaluation set.

**cpg0000** [3]

cpg0000 is a large-scale drug perturbation imaging dataset profiling A549 and U2OS cell lines under more than 300 chemical perturbations. The dataset consists of five-channel fluorescence microscopy images, including Alexa 647, Alexa 568, Alexa 488 long, Alexa 488, and Hoechst 33342. Images are organized across multiple experimental batches, with each batch corresponding to a 384-well plate; each well contains nine imaging sites. In this work, we selected a subset of images from batches BR00116991 and BR00116995 and performed single-cell segmentation using Cellpose, with the Hoechst channel serving as the nuclear reference, to construct the pretraining dataset.

**PBC** [4]

PBC is a curated bright-field single-cell microscopy dataset comprising 17,092 images of peripheral blood cells acquired using a CellaVision DM96 analyzer at the Hospital Clínic of Barcelona. The dataset includes eight clinically relevant cell categories: neutrophils, eosinophils, basophils, lymphocytes, monocytes, immature granulocytes (including promyelocytes, myelocytes, and metamyelocytes), erythroblasts, and platelets. All images were obtained from healthy individuals with no reported infections, hematological disorders, malignancies, or drug treatments, and cell type annotations were provided by expert clinical pathologists.

**HPA** [5]
The Human Protein Atlas (HPA) dataset aims to systematically map protein expression and subcellular localization across human cells and tissues. We used single-cell confocal microscopy images from the HPA subcellular localization subset, where each sample is annotated with cell line, protein localization, and protein identity. To ensure label consistency, we excluded samples associated with multiple protein localizations or protein labels. After filtering, we obtained 579,902 single-cell samples, each represented by four fluorescence channels: microtubules (blue), endoplasmic reticulum (green), DNA (red), and the target protein (alpha). The filtered covers 36 cell lines, 26 subcellular localization categories, and 8,959 proteins, and includes single-cell segmentation annotations, and was split into pretraining and evaluation subsets.

**HepG2** [6]
The HepG2 dataset comprises 520 differential interference contrast (DIC) microscopy images containing 12,198 HepG2 human liver cancer cells, with high-quality ground-truth annotations. A distinctive characteristic of this dataset is the coexistence of multiple cellular states within individual images, capturing both healthy and aberrant adherent cells commonly observed in wet-lab conditions. In this study, only a subset of the segmented single-cell images derived from this dataset was used for pretraining.

**B_ALL** [7]
B_ALL is a quantitative phase imaging (QPI) dataset comprising single-cell images collected from four healthy donors. The dataset includes 710 normal B cells, 389 REH cells, 406 RS4;11 cells, 394 BALL-1 cells, and 415 MN60 cells, providing phase-based morphological profiles for both normal and leukemic B-cell populations.

**Evaluation datasets**
All evaluation datasets were curated to ensure no overlap with the pretraining data. Full-field images were processed using the same preprocessing pipeline as in pretraining to maintain consistency, except for DIBaS, which contains bacterial images, and CM4AI, which includes only a limited number of cells; neither dataset underwent single-cell segmentation. Detailed descriptions of each dataset, along with a comprehensive summary of the evaluation data statistics, are provided in Table S2.

**AINU** [8]
The AINU dataset consists of super-resolution microscopy images that capture the spatial organization of core histone H3, RNA polymerase II, or DNA within cell nuclei. It includes images from human somatic cells and human induced pluripotent stem cells (hiPSCs). By combining super-resolution imaging with detailed nuclear structural information, AINU enables fine-grained characterization of cellular states and heterogeneity. The evaluation task requires models to predict cell types.

**AIRFIHA** [9]
AIRFIHA is a quantitative phase imaging microscopy dataset acquired using diffraction phase microscopy (DPM). It contains phase images of labeled white blood cells collected from multiple donors, including monocytes, granulocytes, B lymphocytes, T lymphocytes, CD4 cells, and CD8 cells. The evaluation task requires models to predict cell types.

**BBBC014** [10]
BBBC014 is a fluorescence microscopy dataset acquired at 10× magnification using a CellCard reader (Vitra Bioscience). Each well contains a single field of view with two images: a nuclear counterstain (DAPI) channel and a signal stain (FITC) channel. The dataset captures nuclear translocation of the transcription factor NFκB in MCF7 (human breast cancer) and A549 (human lung epithelial) cells under varying TNFα stimulation levels. The evaluation task is to predict the cell line identity.

**BBBC021** [11]
BBBC021 consists of fluorescence microscopy images of MCF-7 breast cancer cells treated with 113 small-molecule compounds at eight concentrations for 24 hours. Cells were fixed and stained for DNA, F-actin, and β-tubulin, yielding three-channel fluorescence images. This dataset is designed to benchmark image-based phenotypic profiling methods for drug perturbation analysis. The evaluation task requires models to infer perturbing compounds or their mechanisms of action from image-derived features.

**BBBC048** [12]
BBBC048 contains bright-field images and two fluorescence channels acquired using the ImageStream platform. The dataset includes images of 32,266 asynchronously growing Jurkat cells stained with propidium iodide (PI) to quantify DNA content and MPM2 antibody to identify mitotic cells. Cells were annotated into 7 cell-cycle stages: G1, S, G2, prophase, metaphase, anaphase, and telophase. The evaluation task is prediction of cell-cycle stage.

**CM4AI** [13]
CM4AI is a confocal microscopy dataset generated by the Bridge2AI Cell Map for AI (CM4AI) project. Images contain four channels: nucleus (DAPI, blue), microtubules (red), endoplasmic reticulum (yellow), and a target protein (green). The dataset aims to characterize subcellular structural changes in triple-negative breast cancer cells treated with the anticancer drugs Vorinostat or Paclitaxel. The evaluation task is treatment type classification.

**CNMC** [14]
CNMC is a bright-field microscopy dataset derived from a subset of the C-NMC 2019 dataset. It contains 12,528 white blood cell images from 118 patients, including 8,491 cancerous cells and 4,037 normal cells. All images were annotated by expert pathologists. The evaluation task is cell type classification.

**CoNSeP** [15]
CoNSeP consists of bright-field microscopy images cropped from 41 H&E-stained whole-slide images using provided segmentation masks. Individual cells were extracted and padded when necessary to achieve uniform image size. The dataset includes 4 cell categories: other, inflammatory, epithelial, and spindle-shaped cells. The evaluation task is cell type classification.

**COOS7** [16]
COOS7 is a single-cell fluorescence microscopy dataset consisting of five subsets acquired under systematically varied experimental conditions to assess robustness against technical and temporal variation. Subset 1 contains images collected from four independent culture plates under standard acquisition conditions. Subset 2 comprises images randomly sampled from the same culture plates as Subset 1. Subset 3 includes images acquired from different wells within the same plates. Subset 4 contains images generated several months after the initial acquisition, capturing temporal variation. Subset 5 consists of images acquired using a different imaging instrument, introducing hardware-related variability. Each image is a cropped field centered on a single mouse cell and contains two fluorescence channels: one labeling a target protein associated with a specific subcellular compartment, and one labeling the nucleus. The dataset covers 7 protein localization categories: endoplasmic reticulum, mitochondrial inner membrane, Golgi apparatus, peroxisome, early endosome, cytosol, and nuclear membrane. The evaluation task is protein subcellular localization prediction.

**CYCLoPs** [16]
CYCLoPs is a single-cell fluorescence microscopy dataset of Saccharomyces cerevisiae (yeast) cells, where each image corresponds to a cropped region centered on an individual cell. Images contain two channels, with one channel labeling the target protein and the other labeling the cytosol. After merging the spindle pole category into the spindle class, the dataset includes 16 protein localization categories: actin, bud neck, bud tip, cell periphery, cytoplasm, endoplasmic reticulum, Golgi apparatus, mitochondria, nucleus, nucleolus, peroxisome, spindle, vacuole, vacuolar membrane, cytoplasmic puncta, and cell cortex. The evaluation task is protein localization classification.

**DIBaS** [17]
DIBaS is a bright-field microscopy dataset containing images of 33 bacterial species. The evaluation task is bacterial species classification.

**FCD** [18]
The FCD dataset consists of differential interference contrast (DIC) microscopy images related to hemoglobinopathies. It focuses on a 500 bp motif upstream of the human γ-globin gene locus, where CRISPR-mediated deletion reactivates γ-globin expression.

The dataset includes two cell types, FCD-WT and FCD-HT, and the evaluation task is cell type classification.

**Raabin** [19]
Raabin is a bright-field microscopy dataset acquired from approximately 73 peripheral blood smear images. It includes images of neutrophils, eosinophils, basophils, lymphocytes, and monocytes. Due to the low abundance of basophils in normal samples, additional basophil images were obtained from a chronic myelogenous leukemia (CML) sample. All images were annotated by experts, and the evaluation task is cell type classification.

**Model architecture**
scMIR is built upon the first-stage multimodal architecture of BLIP-2 [20], which consists of a frozen image encoder and a trainable Querying Transformer (Q-Former) that bridges visual representations and language supervision. We initialized the Q-Former using publicly released BLIP-2 pretrained weights and retained its original architecture and parameterization. The image encoder follows the standard BLIP-2 configuration and remains unchanged throughout this work.

On top of the BLIP-2 backbone, we introduce an image reconstruction module that augments the Q-Former with an additional self-supervised learning signal. This module is designed to encourage the Q-Former to preserve fine-grained visual information while learning semantically aligned representations. Importantly, the reconstruction pathway shares parameters with the image–text alignment Q-Former, ensuring that both objectives jointly shape a unified representation space.

Specifically, given an input microscopy image, the frozen image encoder first produces a sequence of visual token embeddings. A subset of these visual tokens is then randomly masked with a fixed masking ratio, while the remaining tokens are retained as contextual inputs. The masked visual token sequence is passed to the Q-Former together with a set of learnable query tokens, following the same cross-attention mechanism used in the image–text alignment stage. Through this shared Q-Former, the query tokens extract latent representations from the partially observed visual features.

To reconstruct the original visual tokens, the masked visual embeddings are projected into the Q-Former hidden space and used as memory inputs to a lightweight Transformer decoder composed of two decoder layers. The decoder takes the Q-Former query outputs as target sequences and attends to the projected visual memory to predict the masked visual token representations. A linear projection layer maps the decoder outputs back to the original visual feature dimension. Reconstruction loss is computed only on masked tokens, using a smooth L1 loss between the predicted and original visual embeddings. The reconstruction objective is weighted by a scalar coefficient and jointly optimized with the image–text alignment objectives during pretraining.

By integrating masked image reconstruction into the BLIP-2 Q-Former framework, scMIR explicitly enforces visual fidelity while preserving the semantic structure induced by multimodal supervision. A complete description of the original BLIP-2 architecture and pretraining objectives can be found in the corresponding reference.

### Pre-training protocol

Pretraining was performed using the AdamW optimizer on an NVIDIA A100-SXM4-80GB GPU under a Linux environment.

## BENCHMARKING

**CytoImageNet** [16]

CytoImageNet is a large-scale microscopy image pretraining dataset inspired by ImageNet, curated to support representation learning for biological imaging. It comprises approximately 890,000 openly sourced and weakly labeled microscopy images spanning 894 classes. Models pretrained on CytoImageNet have been shown to produce feature representations that are competitive with ImageNet-pretrained models on downstream microscopy classification tasks, demonstrating the effectiveness of domain-specific pretraining for biological images. In this study, we adopted the publicly released EfficientNet-B0 backbone pretrained on CytoImageNet, using the official weights provided by the dataset authors.

**EfficientNetB0** [21]

EfficientNet-B0 is a convolutional neural network architecture derived from a systematic model scaling strategy that jointly balances network depth, width, and input resolution using a compound scaling coefficient. This design enables EfficientNet models to achieve strong accuracy–efficiency trade-offs across a wide range of vision tasks. EfficientNet-B0 serves as the base model of the EfficientNet family and is widely used as a standard feature extractor in transfer learning settings. In our experiments, EfficientNet-B0 was initialized with ImageNet-pretrained weights and implemented using the tensorflow.keras.applications package, following standard practice. This model is also included as a reference baseline in the original CytoImageNet benchmark.

**Microsnoop** [22]

Microsnoop is a deep learning–based general-purpose microscopy image representation model trained using masked self-supervised learning on large-scale and heterogeneous microscopy datasets. It is designed to handle diverse imaging scenarios, including single-cell images, full-field microscopy images, and batch-level experimental data. Microsnoop demonstrates robust and state-of-the-art performance across multiple microscopy benchmarks, outperforming both generalists pretrained models and several task-specific approaches. In this work, we used the publicly released Microsnoop pretrained model as a frozen feature extractor, following the recommended inference pipeline provided by the authors.

# EXPERIMENTAL DETAILS

## Cell classification

In addition to general-purpose pretrained models, we included task-oriented baseline methods that are specifically designed for individual datasets and tasks. For each dataset, we either reproduced the feature extraction and classification pipeline described in the original publication or, when such details were unavailable, adopted ResNet-18 as a standard backbone for feature extraction. Task-oriented models were trained in an end-to-end manner by directly attaching a task-specific classification head to the feature extractor.

For fair comparison, dataset-specific train–test splits were applied consistently across all methods, with the number of training and testing samples following the original dataset design. For several high-quality datasets, strong classification performance can be achieved with relatively limited training data. In these cases, we deliberately reduced the size of the training set to evaluate model robustness under data-scarce or challenging conditions, which better reflects practical application scenarios. Accordingly, our experiments were not intended to exhaustively optimize each method for peak performance on every dataset, but rather to assess their relative behavior and generalization under constrained supervision.

For all classification tasks, models were trained using five-fold cross-validation on the training set. The best-performing model from each fold was selected based on validation performance, resulting in five trained models per dataset. These models were then evaluated independently on the held-out test set, and the final performance was reported as the average across the five runs. For most datasets, folds were constructed by stratified splitting based on cell labels to ensure consistent label proportions across folds. For the CM4AI, folds were defined by grouping samples according to their target proteins, which leads to unequal label distributions across folds but prevents information leakage between proteins.

In contrast to task-oriented approaches, scMIR and other general-purpose pretrained models were evaluated using fixed image-level representations, followed by a lightweight classifier consisting of a single hidden layer (128 units) and a classification layer. All methods were trained using the same learning rate and maximum number of epochs for each dataset. Detailed configurations, including dataset-specific train–test splits, task-oriented feature extractors, learning rates, and training epochs, are summarized in Supplementary Table S3.

## FIGURES

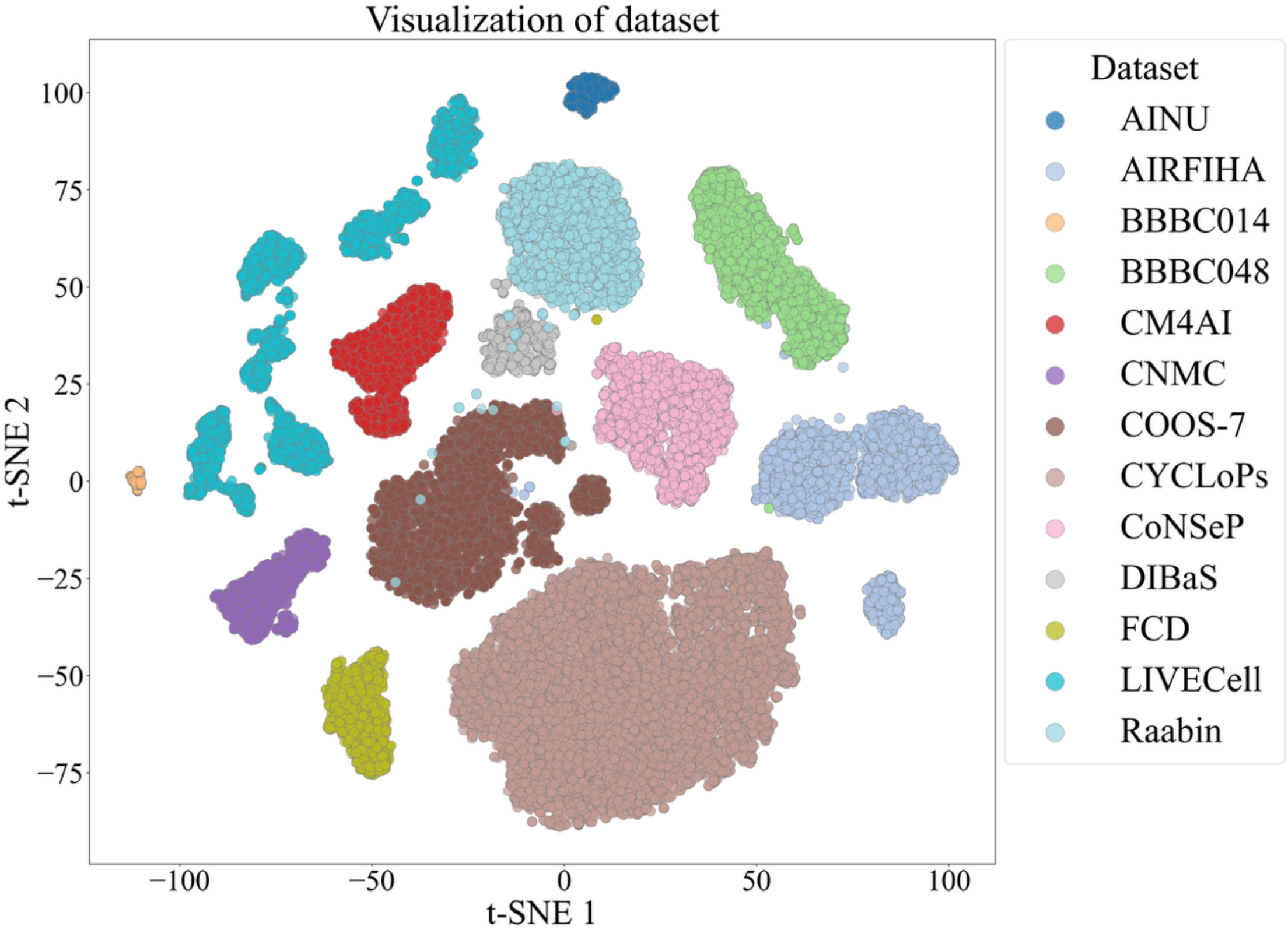

Fig. S1 | t-SNE visualization of CytoImageNet embeddings. Embeddings colored by dataset identity.

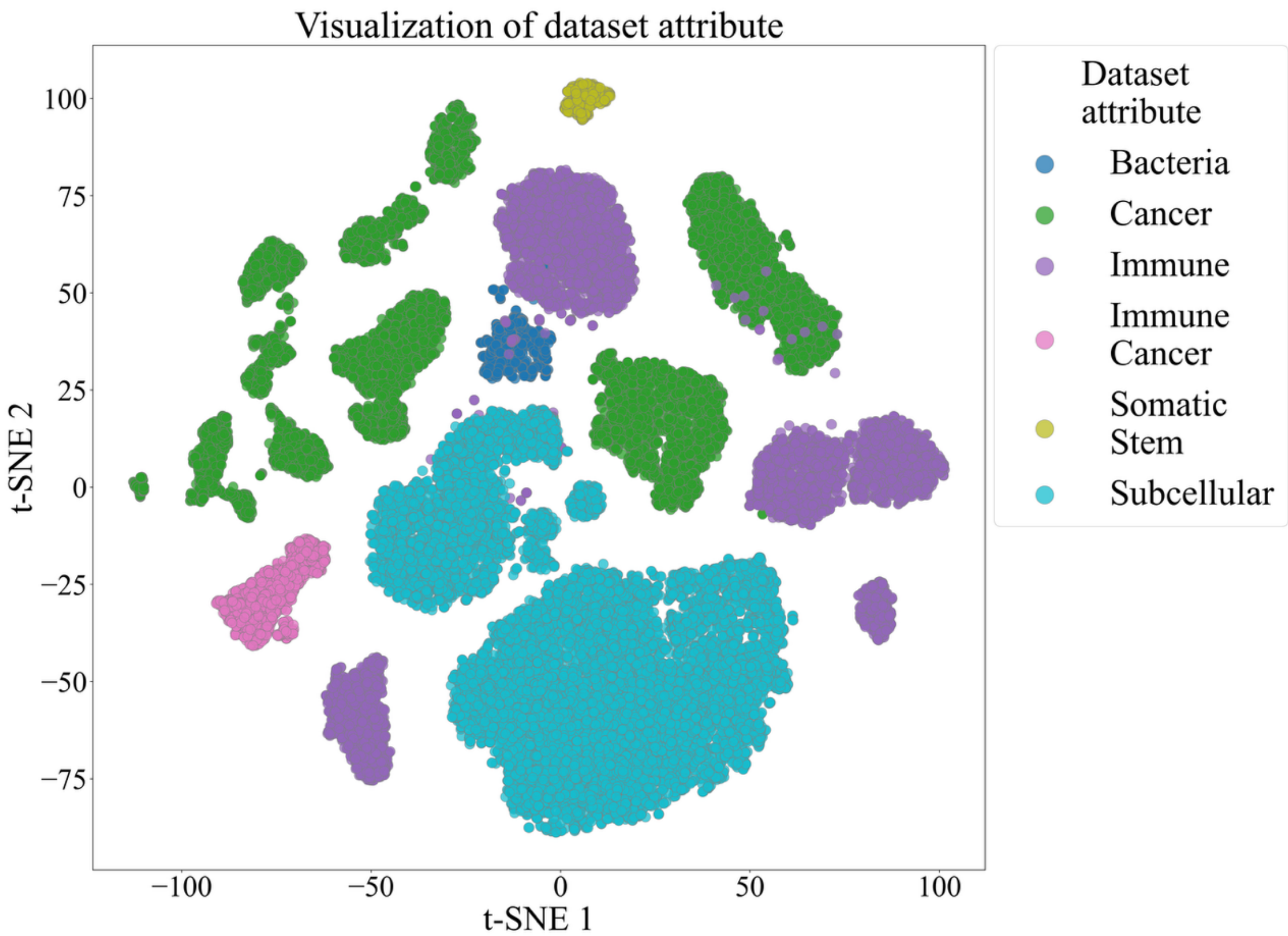

Fig. S2 | t-SNE visualization of CytoImageNet embeddings. Embeddings colored by dataset-level biological attributes.

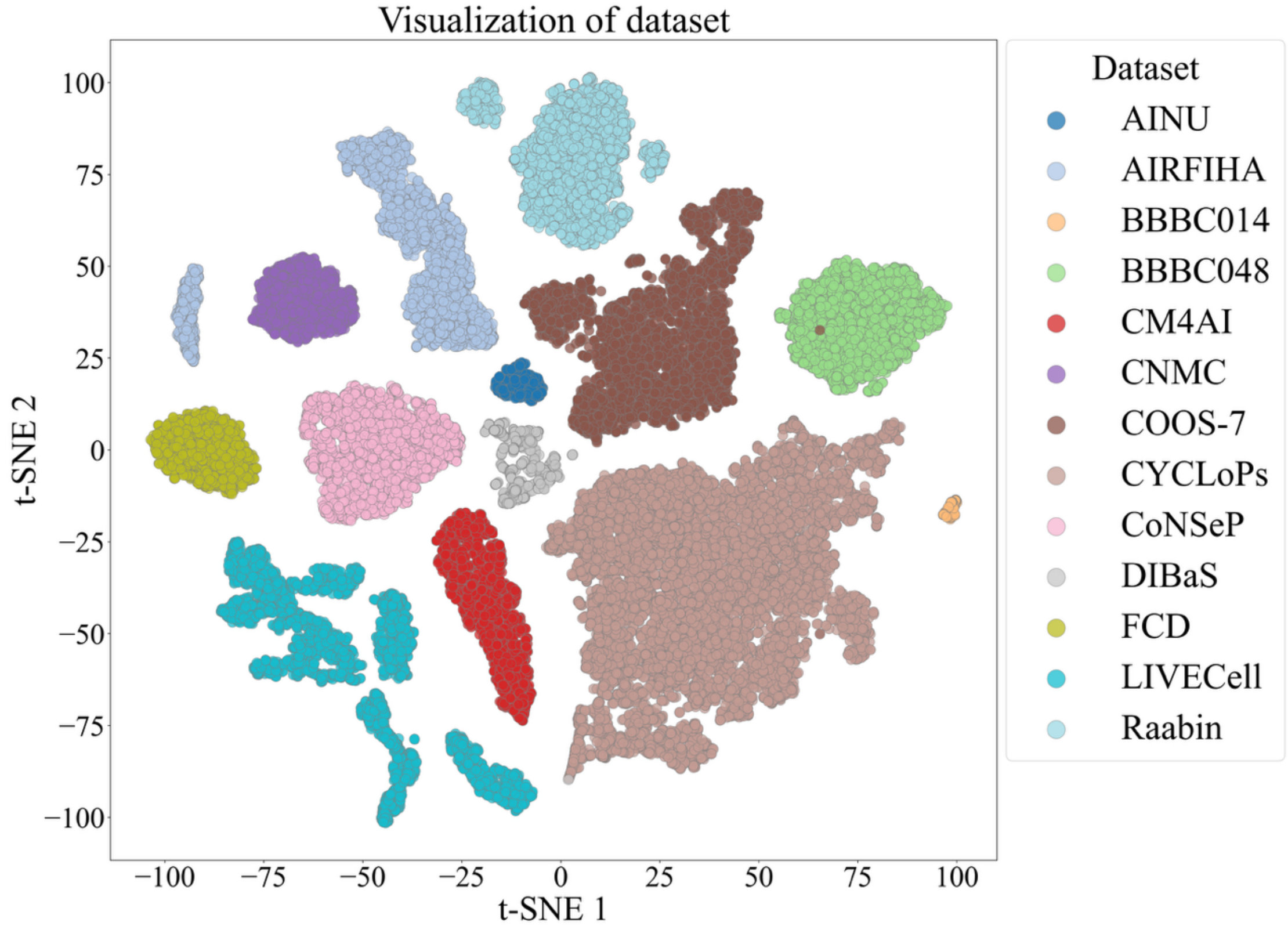

Fig. S3 | t-SNE visualization of EfficientNetB0 embeddings. Embeddings colored by dataset identity.

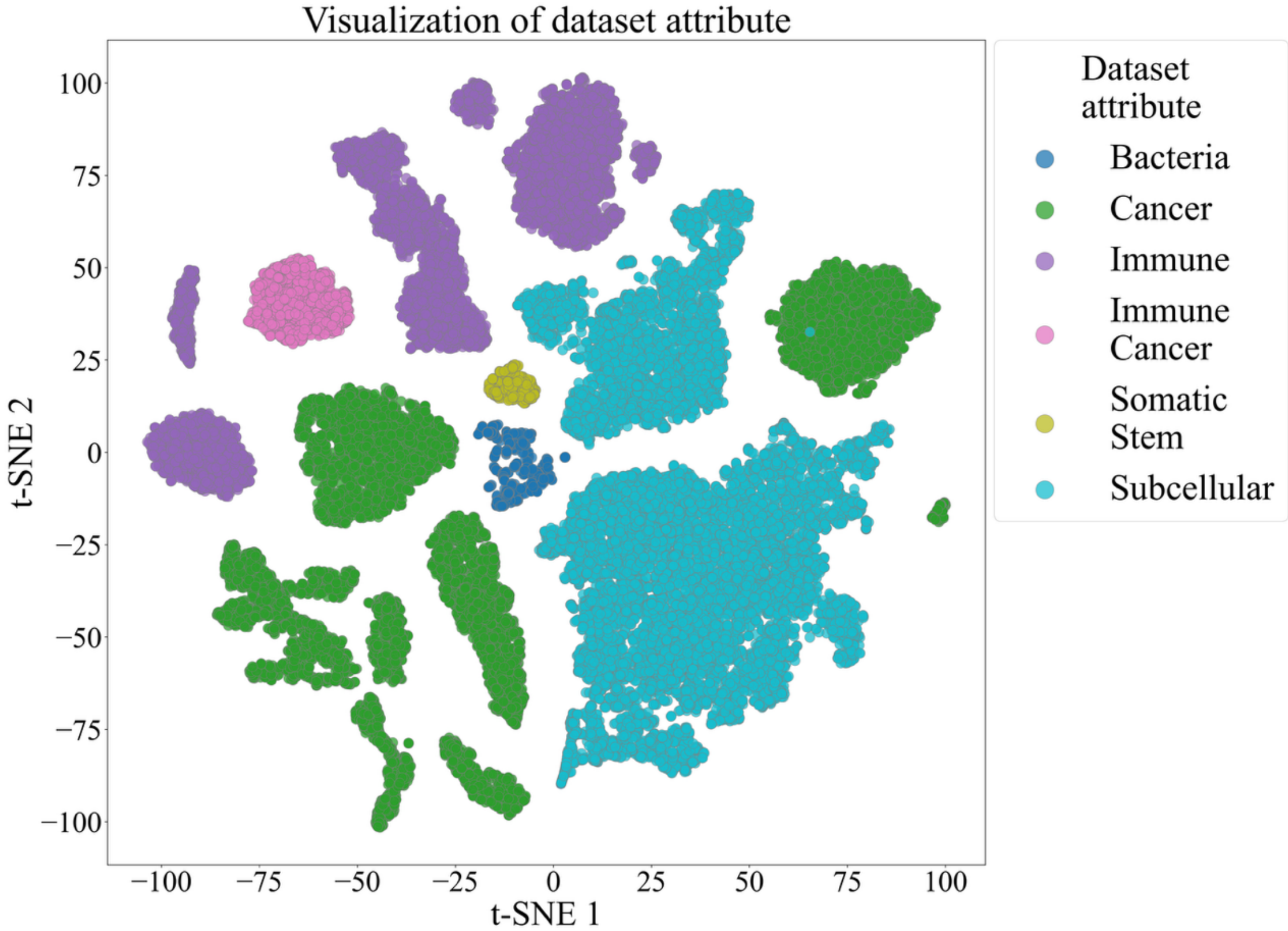

Fig. S4 | t-SNE visualization of EfficientNetB0 embeddings. Embeddings colored by dataset-level biological attributes.

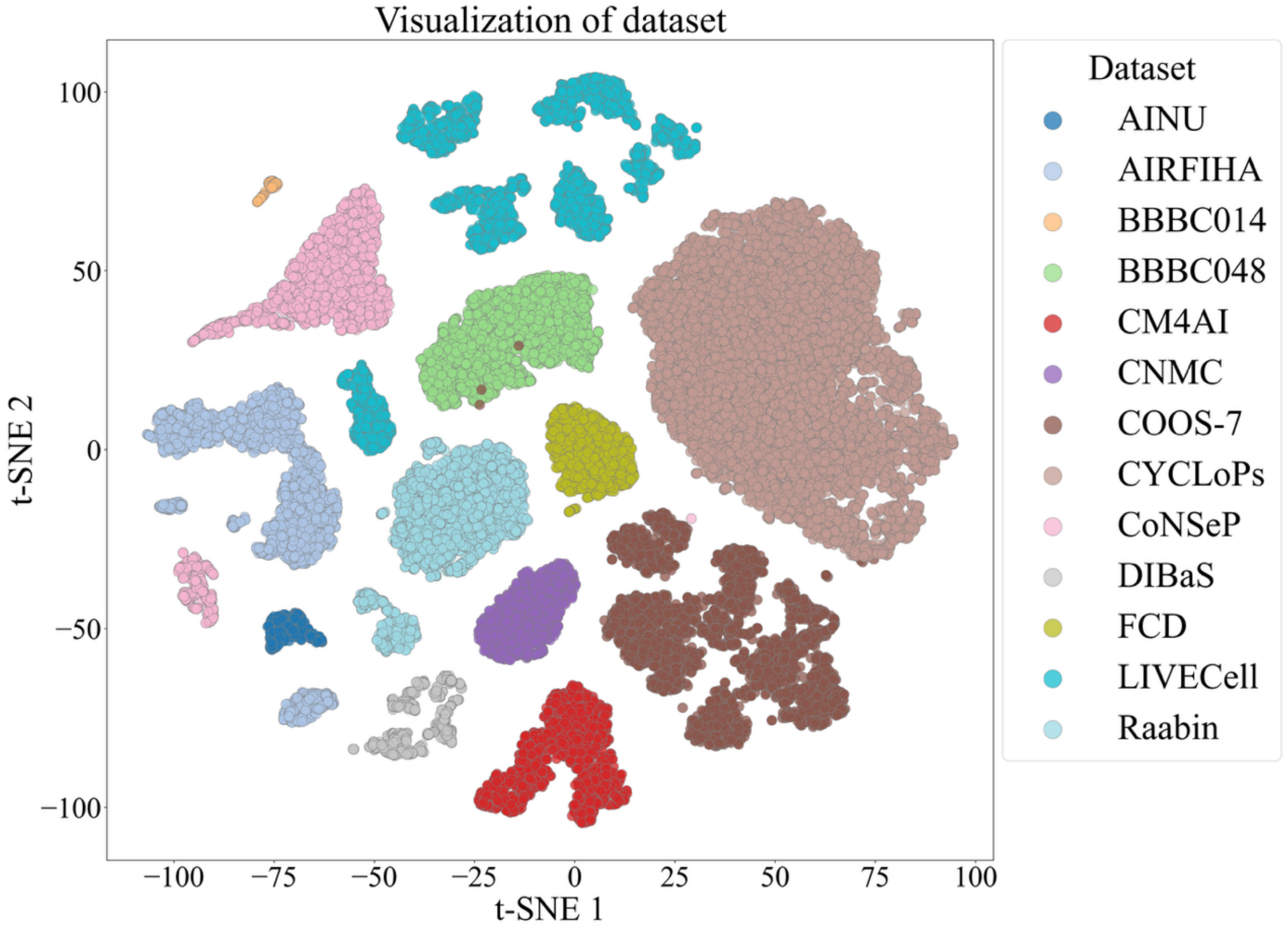

Fig. S5 | t-SNE visualization of Microsnoop embeddings. Embeddings colored by dataset identity.

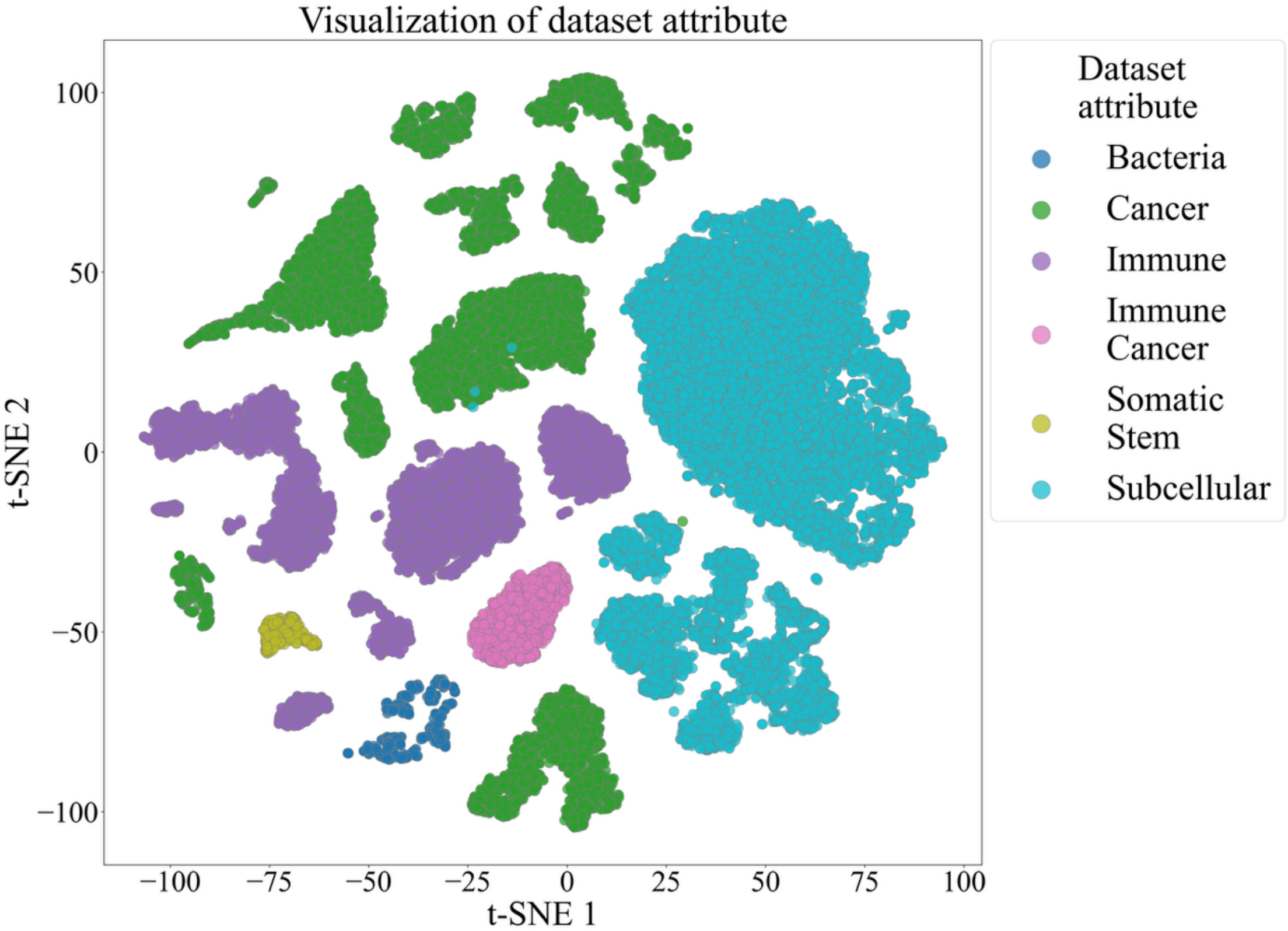

Fig. S6 | t-SNE visualization of Microsnoop embeddings. Embeddings colored by dataset-level biological attributes.

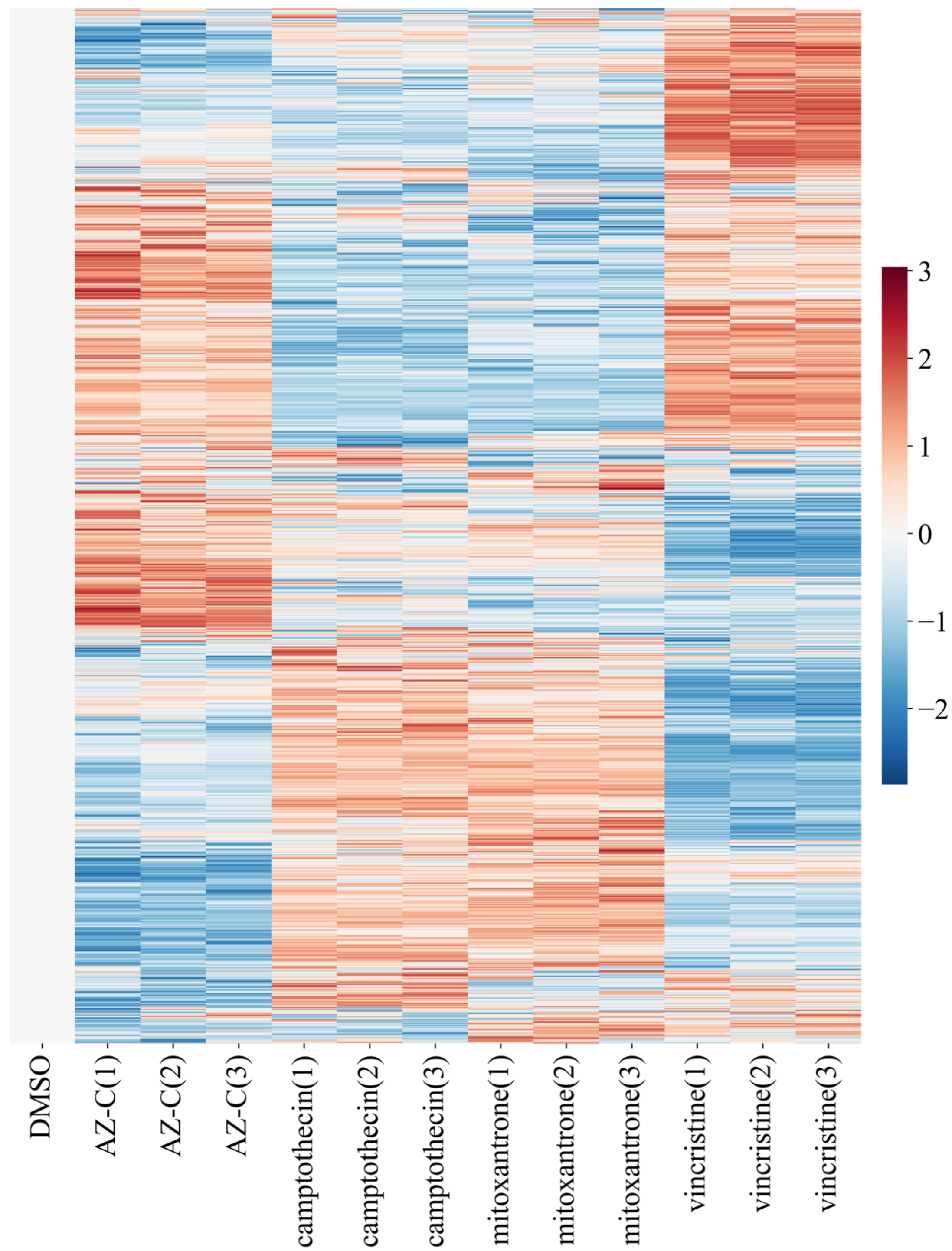

Fig. S7 | Heatmap visualization of different perturbation representations at a concentration of 0.003 on BBBC021.

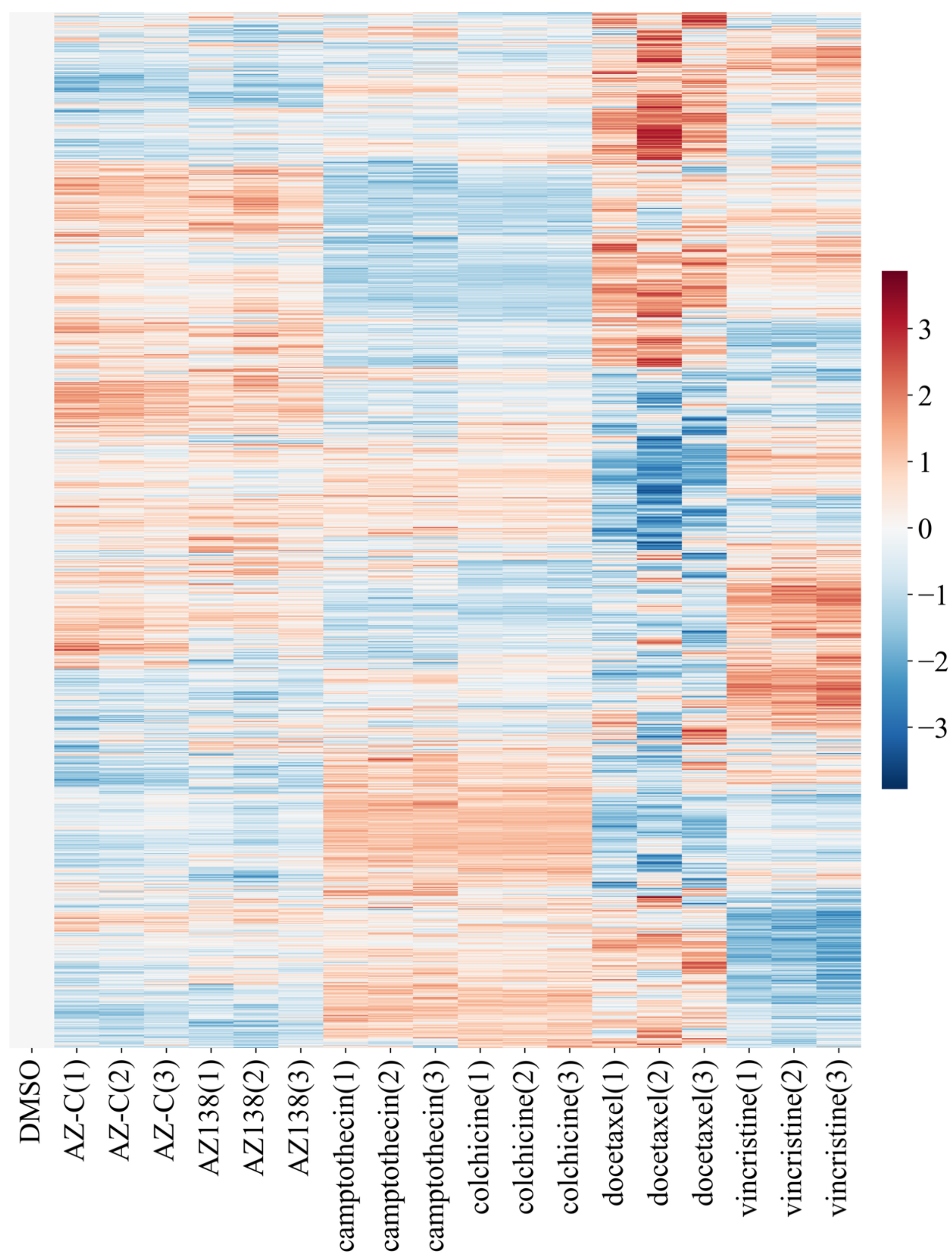

Fig. S8 | Heatmap visualization of different perturbation representations at a concentration of 0.03 on BBBC021.

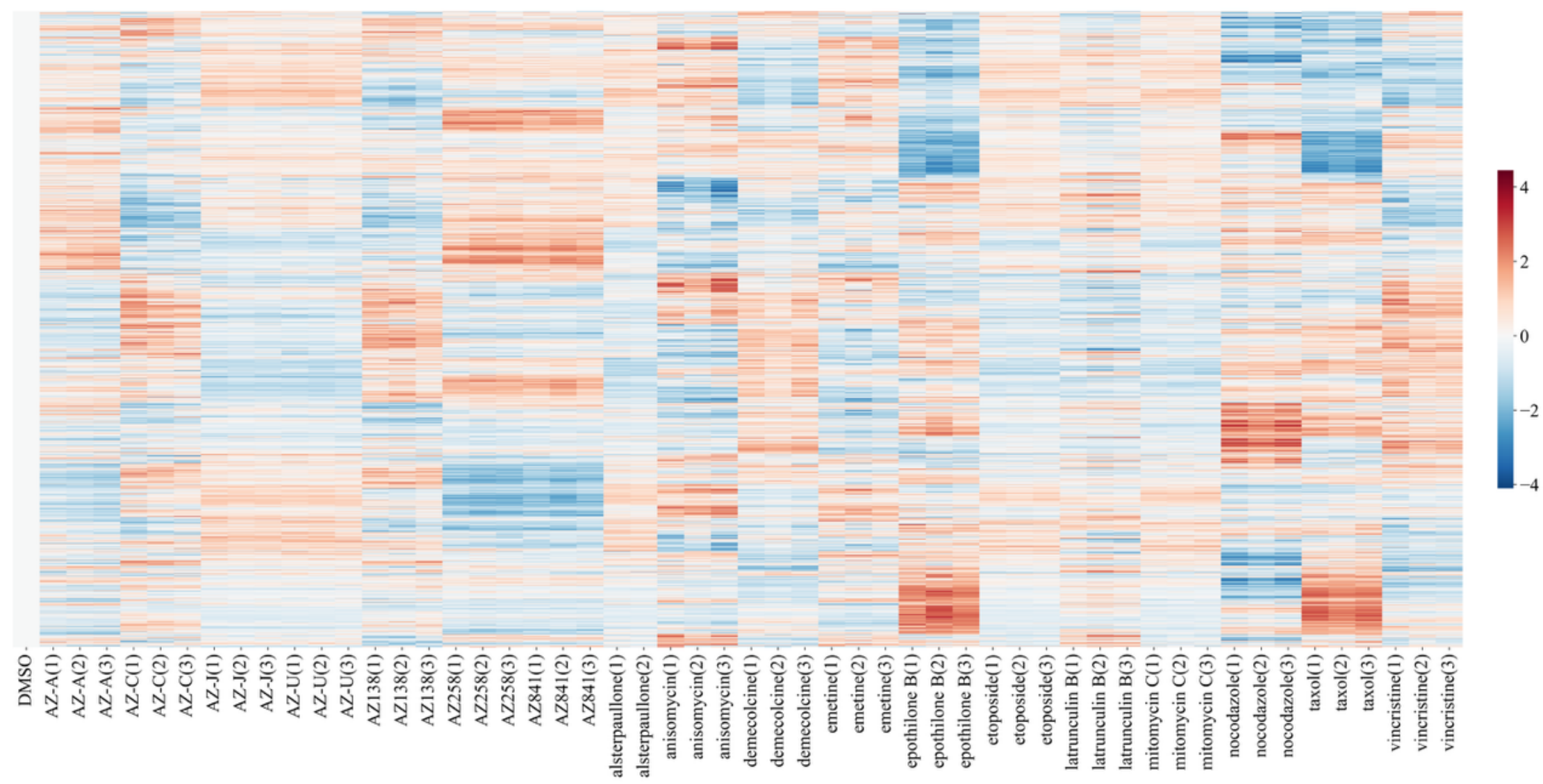

Fig. S9 | Heatmap visualization of different perturbation representations at a concentration of 1.0 on BBBC021.

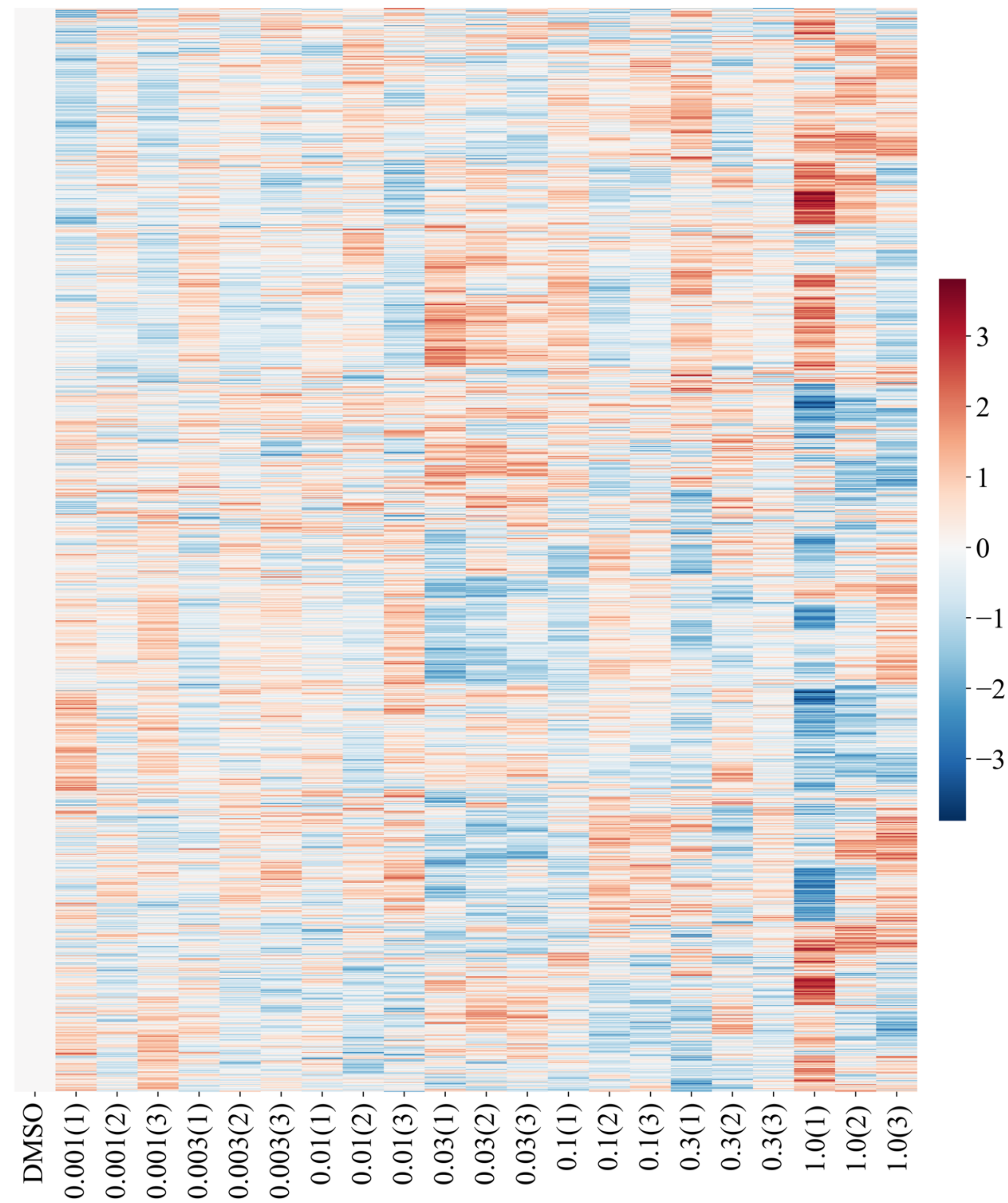

Fig. S10 | Heatmap visualization of of AZ-C perturbation representations.

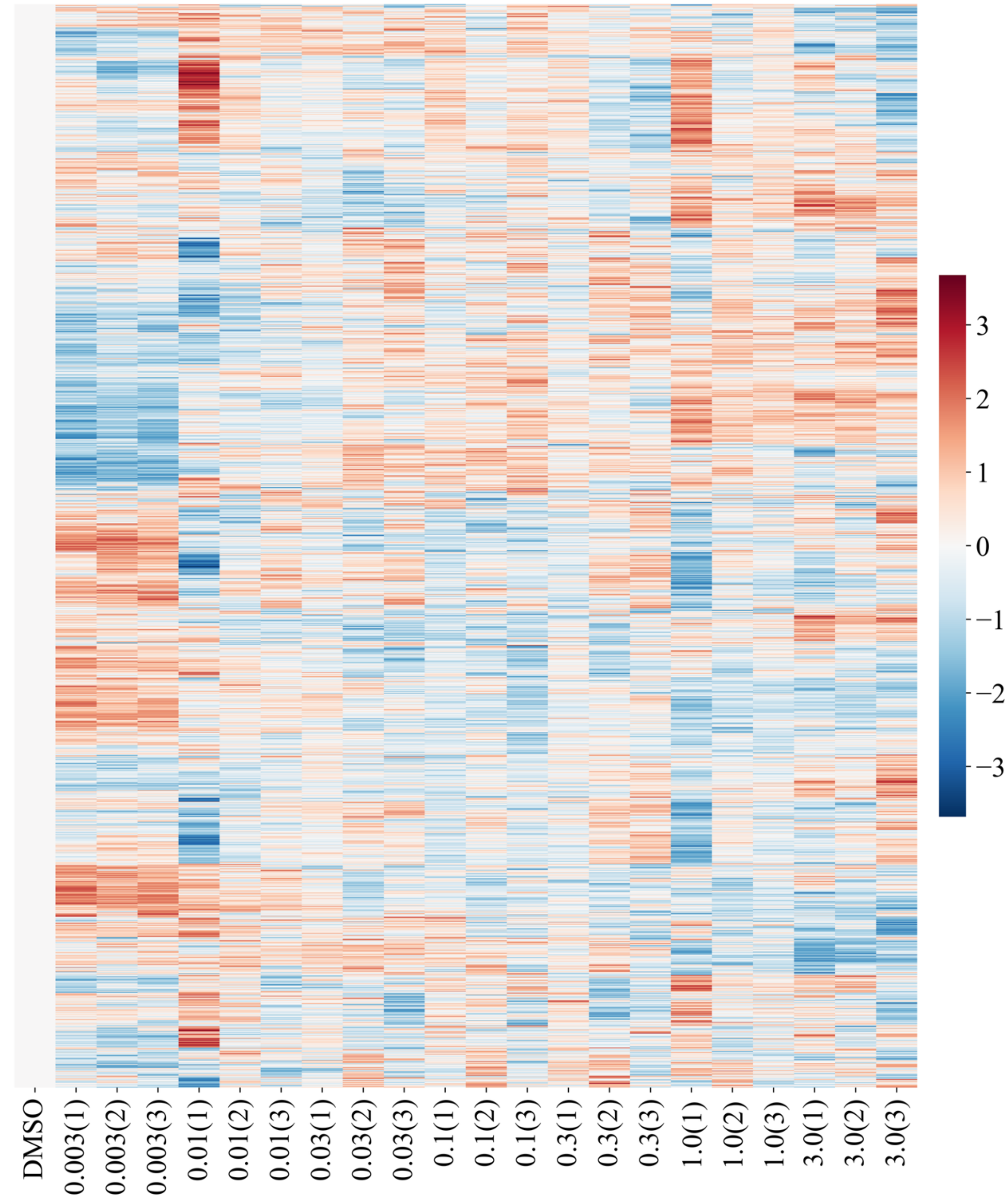

Fig. S11 | Heatmap visualization of of AZ-C perturbation representations.

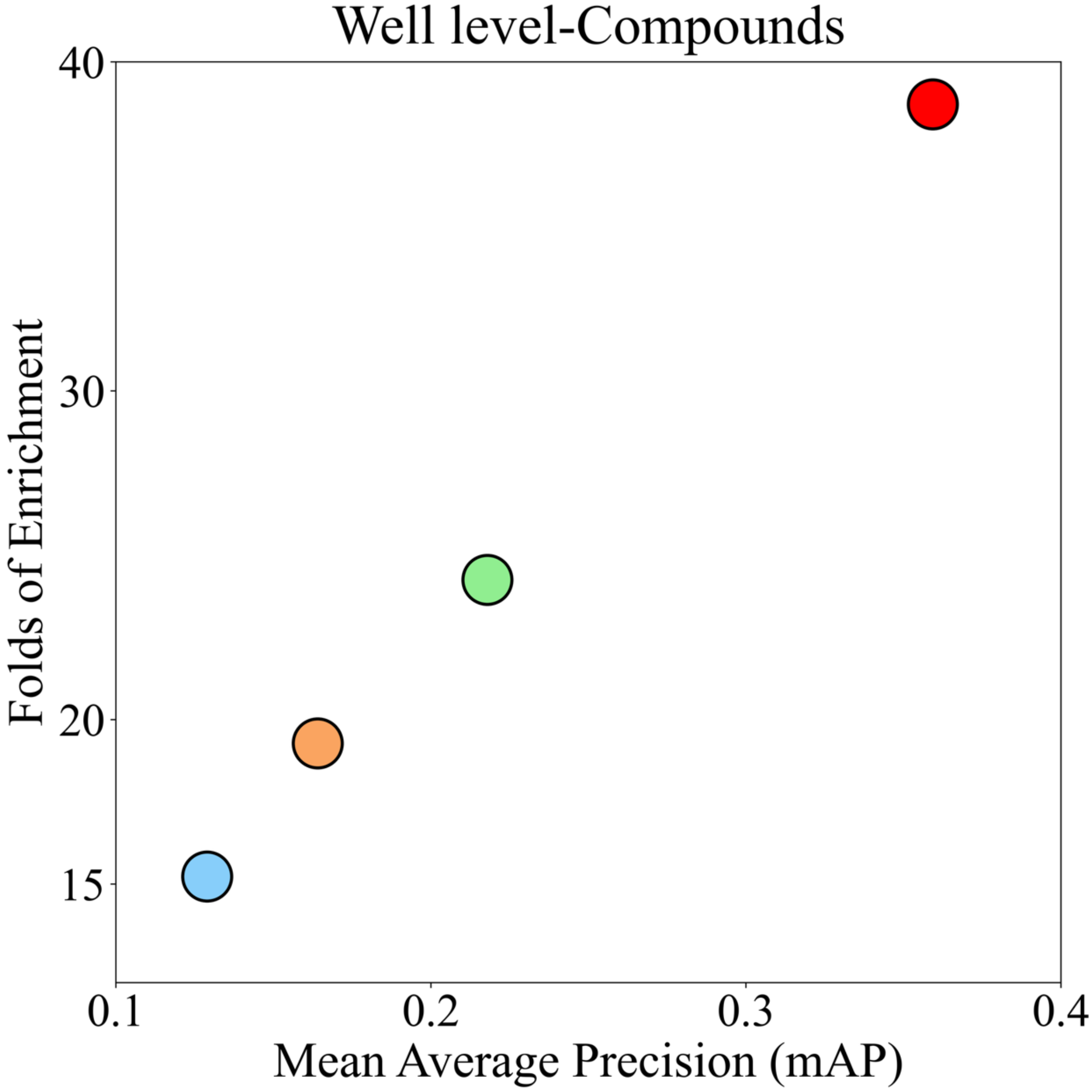

Fig. S12 | Performance comparison on cpg0000.

# TABLES

Table S1. Summary of the pre-training set.

| Datasets Name | Image attribute shape | Image format | Microscopy type | Image number | Biological attributes | Source data download URL |
|---|---|---|---|---|---|---|
| LIVECell | Single-cell MIXED | tif | phase-contrast | 18,632 | multiple cell lines | https://sartorius-research.github.io/LIVECell/ |
| cpg0000 | Single-cell MIXED | tiff | widefield fluorescence | 95,125 | compound perturbation | https://github.com/jump-cellpainting/2024_Chandrasekaran_NatureMethods_CPJUMP1 |
| PBC | Single-cell 360×363×3 | jpg | bright-field | 17,092 | blood cells | https://www.kaggle.com/datasets/orvile/microscopic-peripheral-blood-cell-images |
| HPA | Single-cell MIXED | tif | confocal | 72,832 | protein localization | https://virtualcellmodels.cziscience.com/dataset/hpa-subcellular-section-subcell |
| HepG2 | Single-cell MIXED | png | differential interference contrast | 1,962 | liver cancer | https://zenodo.org/records/13120679 |
| B_ALL | Single-cell 801×801×1 | tiff | quantitative phase imaging | 2,314 | leukemia | https://github.com/vayyappan/QPI_DeepLearn |

Table S2. Summary of the evaluation set.

| Datasets Name | Image attribute shape | Image format | Microscopy type | Image number | Biological attributes | Source data download URL |
|---|---|---|---|---|---|---|
| AINU | Single-cell 1240×1050×3 | png | super-resolution | 349 | cell types | https://doi.org/10.24433/CO.7405455.v2 |
| AIRFIHA | Single-cell 300×300×1 | png | quantitative phase imaging | 3,673 | blood cells | Image data are provided by paper [9] |
| BBBC014 | Full-field 1360×1024×2 | bmp | widefield fluorescence | 96 | cell lines | https://bbbc.broadinstitute.org/BBBC014 |
| BBBC021 | Full-field 1280×1024×3 | tif | widefield fluorescence | 39,600 | compound perturbation | https://bbbc.broadinstitute.org/BBBC021 |
| BBBC048 | Single-cell 66×66×3 | jpg | bright-field +widefield fluorescence | 96,798 | jurkat cell | https://bbbc.broadinstitute.org/BBBC048 |
| CM4AI | Full-field 2048×2048×4 | tif | confocal | 8,908 | compound perturbation | https://dataverse.lib.virginia.edu/dataset.xhtml?persistentId=doi:10.18130/V3/DXWOS5 |
| CNMC | Single-cell 450×450×3 | bmp | bright-field | 12,528 | white blood cell | https://www.kaggle.com/datasets/shafiullahshafin/c-nmc-2019-dataset |
| CoNSeP | Single-cell 112×112×1 | tif | bright-field | 24,332 | cancer cell | https://www.kaggle.com/datasets/karthikperupogu/consep |
| COOS7 | Single-cell 64×64×2 | tif | widefield fluorescence | 264,418 | mouse cell | https://www.kaggle.com/datasets/stanleyhua/coos-7 |
| cpg0000 | Full-field 1080×1080×5 | tiff | widefield fluorescence | 69,120 | compound perturbation | https://github.com/jump-cellpainting/2024_Chandrasekaran_NatureMethods_CPJUMP1 |
| CYCLoPs | Single-cell 64×64×2 | tif | confocal | 56,332 | yeast | https://www.kaggle.com/datasets/stanleyhua/cyclops-protein-loc |
| DIBaS | Full-field 2048×1532×1 | tif | bright-field | 660 | bacteria | https://github.com/gallardorafael/DIBaS-Dataset |
| FCD | Single-cell 100×100×1 | png | differential interference contrast | 3,289 | cell types | Image data are provided by paper [18] |
| HPA | Single-cell 1024×1024×4 | tif | confocal | 2,301,400 | protein localization | https://virtualcellmodels.cziscience.com/dataset/hpa-subcellular-section-subcell |
| LIVECell | Full-field 704×520×1 | tif | phase-contrast | 4,888 | multiple cell lines | https://sartorius-research.github.io/LIVECell/ |
| Raabin | Single-cell 575×575×3 | jpg | bright-field | 16,634 | white blood cell | https://raabindata.com/free-data/ |

Table S3. Summary of the classification task details.

| Datasets Name | Feature extractor | Number of training sample | Number of test sample | Learning rates | Max epochs |
|---|---|---|---|---|---|
| AINU | DenseNet-121 | 223 | 126 | 1e-4 | 1000 |
| AIRFIHA | ResNet-10 | 3,188 | 486 | 1e-4 | 1000 |
| BBBC014 | ResNet-18 | 57 | 37 | 1e-4 | 300 |
| BBBC048 | ResNet-18 | 19,359 | 12,907 | 1e-4 | 1000 |
| CM4AI | ResNet-18 | 192 | 2,035 | 1e-3 | 300 |
| CNMC | ResNeXt-CC | 8,768 | 3,760 | 1e-4 | 1000 |
| CoNSeP | ResNet-18 | 14,599 | 9,733 | 1e-4 | 1000 |
| COOS7 | ResNet-18 | 448 | 41,008<br>10,364<br>17,021<br>32,596<br>30,772 | 1e-4 | 300 |
| CYCLoPs | ResNet-18 | 16,899 | 11,267 | 1e-4 | 1000 |
| DIBaS | ResNet-18 | 396 | 264 | 1e-4 | 1000 |
| FCD | T2D5 | 1,973 | 1,316 | 1e-4 | 1000 |
| LIVECell | ResNet-18 | 384 | 4,504 | 1e-3 | 300 |

| Raabin | ResNet-18 | 320 | 16,314 | 1e-3 | 300 |
|---|---|---|---|---|---|

Table S4. Original localization categories in the HPA dataset with grouping annotated by experts.

| Protein localization | Minor group | Middle group | Major group |
|---|---|---|---|
| Nucleoli rim | Nucleoli | Nucleoli | Nucleoli |
| Nucleoli | Nucleoli | Nucleoli | Nucleoli |
| Nucleoli fibrillar center | Nucleoli fibrillar center | Nucleoli | Nucleoli |
| Vesicles | Vesicles | Vesicles | Endomembrane system |
| Peroxisomes | Peroxisomes | Vesicles | Endomembrane system |
| Lysosomes | Lysosomes | Vesicles | Endomembrane system |
| Endosomes | Endosomes | Vesicles | Endomembrane system |
| Lipid droplets | Lipid droplets | Vesicles | Endomembrane system |
| Endoplasmic reticulum | Endoplasmic reticulum | Endoplasmic reticulum | Endomembrane system |
| Golgi apparatus | Golgi apparatus | Golgi apparatus | Endomembrane system |
| Cytoplasmic bodies | Cytosol | Cytosol | Cytosol |
| Cytosol | Cytosol | Cytosol | Cytosol |
| Centriolar satellite | Centrosome | Centrosome | Cytosol |
| Centrosome | Centrosome | Centrosome | Cytoskeleton |
| Microtubule ends | Microtubules | Microtubules | Cytoskeleton |
| Microtubules | Microtubules | Microtubules | Cytoskeleton |
| Intermediate filaments | Intermediate filaments | Intermediate filaments | Cytoskeleton |
| Actin filaments | Actin filaments | Actin filaments | Cytoskeleton |
| Focal adhesion sites | Actin filaments | Actin filaments | Cytoskeleton |
| Cell Junctions | Plasma membrane | Plasma membrane | Plasma membrane |
| Plasma membrane | Plasma membrane | Plasma membrane | Plasma membrane |
| Mitochondria | Mitochondria | Mitochondria | Mitochondria |
| Nuclear membrane | Nuclear membrane | Nuclear membrane | Nucleus |
| Nucleoplasm | Nucleoplasm | Nucleus | Nucleus |
| Nuclear bodies | Nuclear bodies | Nucleus | Nucleus |
| Nuclear speckles | Nuclear speckles | Nucleus | Nucleus |